\documentclass[10pt,journal,compsoc]{IEEEtran}
\usepackage{amssymb}
\usepackage{multirow}
\usepackage{booktabs}
\usepackage{algorithm}
\usepackage{algorithmic}
\usepackage{subfigure}
\usepackage{flushend}
\usepackage{ragged2e}
\usepackage{url}
\usepackage{threeparttable}

\usepackage{amsmath,amsfonts}
\usepackage{algorithmic}
\usepackage{algorithm}
\usepackage{array}
\usepackage[caption=false,font=normalsize,labelfont=sf,textfont=sf]{subfig}
\usepackage{textcomp}
\usepackage{stfloats}
\usepackage{url}
\usepackage{verbatim}
\usepackage{graphicx}
\usepackage{cite}
\usepackage{bbm}
\usepackage{array}
\newcolumntype{R}{>{\color{red}}c}

\usepackage{amsmath}
\title{
Micro-Macro Spatial-Temporal Graph-based  Encoder-Decoder for Map-Constrained Trajectory Recovery
}

\author{Tonglong Wei, Youfang Lin, Yan Lin, Shengnan Guo*, Lan Zhang, Huaiyu Wan
\thanks{~~*Corresponding author: Shengnan~Guo}
\IEEEcompsocitemizethanks{\IEEEcompsocthanksitem T. Wei, Y. Lin, Y. Lin, S. Guo, L. Zhang and H. Wan are with School of Computer and Information Technology, Beijing Jiaotong University, Beijing 100044, China; And the Beijing Key Laboratory of Traffic Data Analysis and Mining, Beijing 100044, China
\protect\\
E-mail: weitonglong@bjtu.edu.cn; yflin@bjtu.edu.cn; ylincs@bjtu.edu.cn; guoshn@bjtu.edu.cn; zhanglan20@bjtu.edu.cn; hywan@bjtu.edu.cn. }
	\thanks{Manuscript received xx xx, xxxx; revised xx xx, xxxx.}
}

\newtheorem{definition}{Definition}

\usepackage{color}

\begin{document}

\IEEEtitleabstractindextext{%
\begin{abstract}
\justifying
Recovering intermediate missing GPS points in a sparse trajectory, while adhering to the constraints of the road network, could offer deep insights into users' moving behaviors in intelligent transportation systems. 
Although recent studies have demonstrated the advantages of achieving map-constrained trajectory recovery via an end-to-end manner, they still face two significant challenges.
\textcolor{black}{Firstly, existing methods are mostly sequence-based models. It is extremely hard for them to comprehensively capture the micro-semantics of individual trajectory, including the information of each GPS point and the movement between two GPS points.} 
\textcolor{black}{Secondly, existing approaches ignore the impact of the macro-semantics, \emph{i.e.}, the road conditions and the people’s shared travel preferences reflected by a group of trajectories.} 
To address the above challenges, we propose a Micro-Macro Spatial-Temporal Graph-based Encoder-Decoder (MM-STGED). 
Specifically, we model each trajectory as a graph to efficiently describe the micro-semantics of trajectory and design a novel message-passing mechanism to learn trajectory representations. 
Additionally, we extract the macro-semantics of trajectories and further incorporate them into a well-designed graph-based decoder to guide trajectory recovery. 
Extensive experiments conducted on sparse trajectories with three different sampling intervals that are respectively constructed from two real-world trajectory datasets demonstrate the superiority of our proposed model. 
\end{abstract}

}

\maketitle
\IEEEdisplaynontitleabstractindextext
\IEEEpeerreviewmaketitle

\IEEEraisesectionheading{\section{Introduction}\label{sec1}}
\IEEEPARstart{A}{} trajectory is a sequence of timestamped locations that describe the movement of users.
\textcolor{black}{With the deployment of GPS-enabled devices in Intelligent Transportation Systems (ITS), massive user trajectory data is collected. 
Mining trajectory data plays an essential role in understanding the travel patterns of individuals and vehicles and supporting various downstream intelligent navigation applications,} e.g., arrival time estimation~\cite{derrow2021eta, chen2022interpreting}, route prediction~\cite{yang2018efficient,wen2021package, wen2022graph2route}, and anomalous trajectory detection~\cite{liu2020online,han2022deeptea}. 

\begin{figure}
	\centering
	\includegraphics[width=1.0\linewidth]{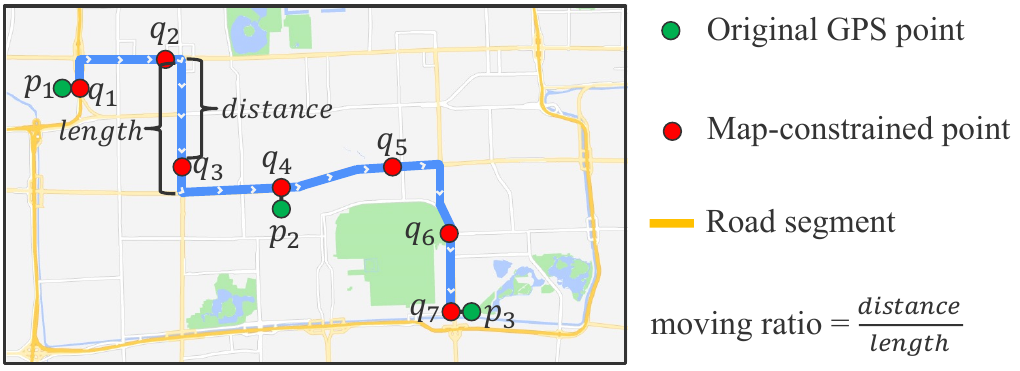}
	\caption{Illustrations of map-constrained trajectory recovery, road segment, and moving ratio.}
	\label{fig:example}
\end{figure}

\textcolor{black}{However, in practice, trajectories are often sparse with a large sampling interval due to the limitation in sampling mechanisms and storage devices. For example, taxis typically report their GPS locations every 2 to 6 minutes to save energy~\cite{yuan2010interactive}. These sparse trajectories often result in a loss of significant information.
Figure~\ref{fig:example} illustrates an example of sparse trajectory consisting of three GPS points with a sampling interval of 90 seconds: $\tau_1 = \langle p_1, p_2, p_3 \rangle$, where $p_i = (lat_i, lng_i, t_i)$ records the location and time information of GPS point, and $t_{i+1} - t_i = 90$ seconds, $i \in \{1, 2\}$.
Due to the large sampling intervals, the distance between GPS points is significant, leading to a loss of information about the intermediate points. Consequently, it is hard to accurately describe the user's detailed movement. 
Furthermore, the large sampling intervals make it difficult to accurately infer the intermediate missing road segments that the user passes by, resulting in losing the road-related information (which road segment the GPS points belong to and the corresponding moving ratio).
However, the road-related information of trajectories is crucial to downstream applications since these applications need to know the relationship between GPS points and roads in transportation networks. 
Therefore, the sparsity of practical trajectories will decrease the performance of downstream trajectory mining tasks.
}



\textcolor{black}{To improve the availability and utilization values of the sparse trajectories, it is necessary to recover the multiple aspects of information on sparse trajectories, including GPS locations and road-related information. 
This kind of task is termed as \textbf{``map-constrained trajectory recovery''}~\cite{ren2021mtrajrec}, which is a crucial trajectory preprocessing step to support downstream trajectory mining tasks.}

\textcolor{black}{Figure~\ref{fig:example} shows an example of the map-constrained trajectory recovery task. 
Given the trajectory $\tau_1 = \langle p_1, p_2, p_3 \rangle$ defined above, 
the map-constrained trajectory recovery task aims to recover a map-constrained trajectory with a sampling interval of 30 seconds: $\tilde{\tau } = \langle q_1, \cdots, q_7 \rangle$, where each map-constrained point $q_j = (e_j, r_j, t_j)$ consists of road segment $e_j$, moving ratio $r_j$, and time $t_j$. And $t_{j+1} - t_j = 30$ seconds, $j \in \{1, \cdots, 6\}$. Meanwhile, the precise GPS location information (latitudes and longitudes) can be calculated by $e$ and $r$.}



\textcolor{black}{Existing works about map-constrained trajectory recovery can be categorized by \textbf{two-stage solutions} and \textbf{end-to-end solutions}.
\textbf{The two-stage solutions }can be further classified into two subcategories: \textit{first recovering the missing GPS points and then performing map matching}, or \textit{first performing map matching and then recovering the map-constrained points}. 
The former pipeline~\cite{hoteit2014estimating,wang2019deep} may suffer from error accumulation, as the recovered GPS points may be inaccurate, leading to unreliable map-matching results. 
The latter pipeline~\cite{jagadeesh2017online, zhao2019deepmm} first maps the sparse GPS points onto the road and then calculates the shortest path between adjacent points. 
Since most map-matching algorithms, which are typically based on the Hidden Markov Model (HMM), struggle to achieve high accuracy with sparse trajectories, this kind of approach usually has limitations in terms of accuracy. 
Additionally, the complex movement patterns of vehicles often deviate from the shortest path.
Moreover, both of the above two-stage solutions are inefficient due to the time-consuming nature of map matching~\cite{newson2009hidden}.}

\textcolor{black}{\textbf{The end-to-end solutions} aim to recover map-contrained points directly.
In comparison to two-stage solutions, end-to-end solutions are more effective as they avoid error accumulation. Additionally, they are more efficient since they eliminate the requirement for map matching.}
However, despite the significant advancements made by recent representative works~\cite{ren2021mtrajrec,chen2023rntrajrec},  end-to-end map-constrained trajectory recovery still faces the following unsolved challenges.

\begin{figure}
	\centering
	\includegraphics[width=0.5\textwidth]{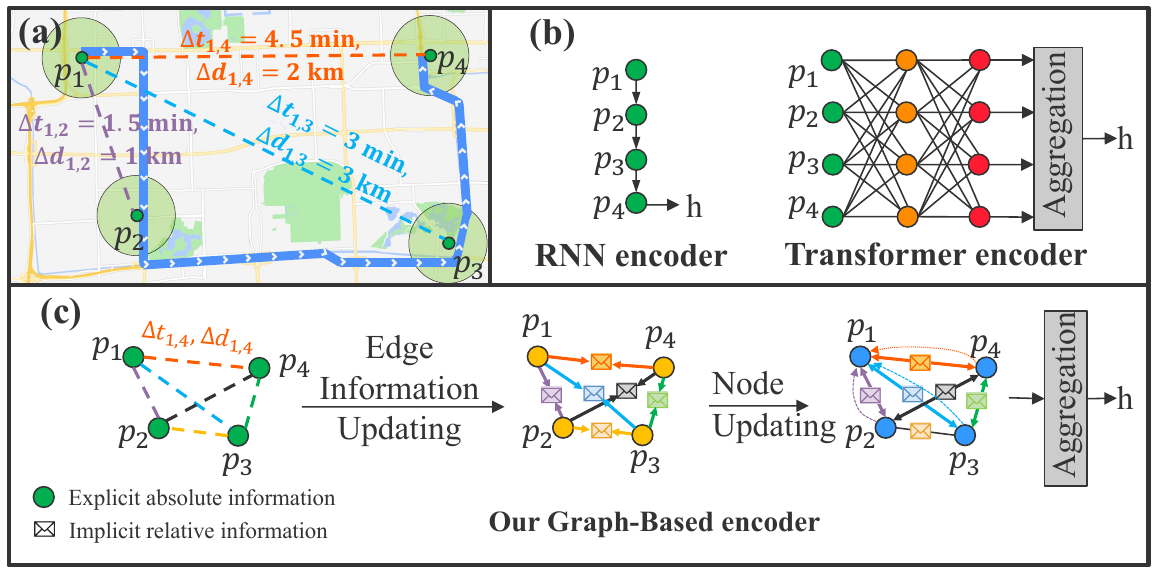}
 \caption{\textcolor{black}{Illustrations of the trajectory’s micro-semantic information and different trajectory encoders. (a) the micro-semantic information of a trajectory $\tau_2 = \langle p_1, \cdots, p_4 \rangle$ includes the explicit absolute information of single GPS points and the implicit relative information between GPS points. (b) Sequence-based encoders. (c) Graph-based encoder.}}
	\label{fig:Motivation_graph}
\end{figure}

\textbf{1) It is extremely hard for existing solutions to fully consider the micro-semantic information of one trajectory, which is important to understand the movement behavior of users.} \textcolor{black}{Specifically, as shown in Figure~\ref{fig:Motivation_graph}(a), for a sparse trajectory $\tau_2 =\langle p_1, p_2, p_3, p_4 \rangle$, its micro-semantic information consists of two parts: }
1) the explicit absolute information provided by each GPS point, including latitude, longitude, timestamp, and surrounding road network, which indicates the discrete positions of the moving objects, and 
2) \textcolor{black}{the implicit relative information between two GPS points, including space transfer $\triangle d_{i,j}$ and time cost $\triangle t_{i,j}$ (e.g., in Figure~\ref{fig:Motivation_graph}(a), $i,j \in \{1,2,3,4\}$ and $i \neq j$), which reflects users' driving behavior between two observed GPS points. This implicit relative information between GPS points describes the continuity of the trajectories.}
\textcolor{black}{For example, based on the time cost, we can infer that the user's movement is first from $p_1$ to $p_3$ then to $p_4$ since $\triangle t_{1,3}<\triangle t_{1,4}$. Additionally, with the space transfer $\triangle d_{1,3}>d_{1,4}$, we can guess the user detours from $p_1$ to $p_4$ instead of following the shortest path.}
Both types of information are valuable in understanding individual trajectories and recovering trajectory details. Therefore, an efficient trajectory learning method should be able to simultaneously capture both types of information. 
However, most existing trajectory recovery methods treat one trajectory as a sequence and adopt sequential models for trajectory learning. As shown in Figure~\ref{fig:Motivation_graph}(b), these models primarily focus on utilizing the explicit absolute information of each GPS point while ignoring the implicit relative information between them. So, the comprehensive capture of micro-semantic information from both aspects has not been adequately addressed by existing related works.

\textcolor{black}{\textbf{2) Fail to \textcolor{black}{explicitly} model the macro-semantic information reflected by a group of trajectories, 
which usually has a significant impact on user individual travel decisions.}
Specifically, the macro-semantic information of the trajectories includes 1) the traffic conditions related to the traveling routes and 2) the people's shared travel preferences. 
The traffic conditions related to the traveling routes represent the environment in which the user is located. 
It is helpful for recovering the details of trajectories since a user's actual driving behavior is affected by the environment. 
For example, if the current route is congested, the user will drive at a slower speed or switch to alternative routes, both of which will affect the trajectory. }
\textcolor{black}{People's shared travel preferences reflect the general route choice habits followed by most people, which can provide prior knowledge to guide the trajectory recovery. However, modeling people's shared travel preferences is a non-trivial task. On the one hand, people's route choices are constrained by the physical topology of the road network, since vehicles must pass on the road. On the other hand, real-time traffic control, road passing rules, and convenience also influence people's route choices, since they must obey traffic rules and strive for a convenient travel experience to their destination. 
For instance, the physically connected road may be impassable due to traffic control. Users might choose routes with longer distances but shorter travel times based on practical considerations like road speed limits and types, rather than opting for the shortest distance routes.
Therefore, how to accurately represent people’s shared travel preferences considering both these two aspects and integrating the traffic conditions related to the traveling routes to guide trajectory recovery deserves thoughtful design.}

To tackle the above challenges and achieve accurate map-constrained trajectory recovery, we propose the ``\emph{\textbf{M}icro-\textbf{M}acro \textbf{S}patial-\textbf{T}emporal \textbf{G}raph-based \textbf{E}ncoder-\textbf{D}ecoder}" (MM-STGED) model. This model simultaneously incorporates the micro-semantics and macro-semantics of trajectories to guide trajectory recovery.
In contrast to existing works that solely focus on utilizing explicit absolute information of each GPS point using a sequential model, MM-STGED models the trajectory from the graph perspective, as shown in Figure~\ref{fig:Motivation_graph}(c). This allows the model to capture both the explicit absolute information of GPS points and their implicit relative information by utilizing the nodes and edges of the graph.
Additionally, MM-STGED introduces a novel graph convolution mechanism to aggregate node and edge information, enabling it to simultaneously capture the discreteness and continuity of trajectories efficiently. 
To extract the macro-semantics from a group of trajectories, we construct a macro trajectory flow graph according to trajectory data and road networks, representing people's shared travel preferences. Furthermore, we propose to describe the traffic conditions only relying on the available trajectory data, eliminating the need for external information. 
Finally, MM-STGED utilizes a graph-based decoder to incorporate macro-semantics into the trajectory recovery task. 
In general, our contributions can be summarized as follows:

\begin{itemize}
    \item We are the first to model trajectories from a graph perspective to perform the map-constrained trajectory recovery task. Specifically, we propose a new framework, namely MM-STGED, which comprehensively captures the micro- and macro-semantics of trajectories.
    \item A graph-based encoder is designed to model trajectory's micro-semantics, which utilizes nodes and edges to describe the explicit absolute information of GPS points and the implicit relative information between them, as well as a novel graph convolution is proposed to simultaneously aggregate them together.
    \item A graph-based decoder is designed to incorporate the trajectories' macro-semantics, which models and utilizes the people's shared travel preferences and traffic conditions to guide trajectory recovery.
    \item Extensive experiments conducted on two real-world trajectory datasets with three different sampling intervals demonstrate that our method significantly outperforms the state-of-the-art baselines. 
\end{itemize}

\section{Related Work}
\subsection{Map-constrained Trajectory Recovery}
Recovering map-constrained trajectories is essential in intelligent transportation systems, and can support various applications, e.g., arrival time estimation \cite{chen2022interpreting}, trajectory prediction \cite{sadri2018will}, trajectory similarity computation \cite{fang2022spatio,zhang2020trajectory,li2021spatial}, and anomalous detection \cite{zhang2011ibat,liu2020online}. 
There are many efforts have been made on this topic.

The traditional solution usually consists of two stages, including \textit{first recovering then performing map matching} and \textit{first performing map matching then recovering}.
\textcolor{black}{\textit{First recovering then performing map matching.} To recover the missing GPS points in a sparse trajectory, several approaches have been proposed~\cite{hoteit2014estimating, elshrif2022network, wang2019deep, chen2018tripimputor, li2021trajectory}. For instance, Linear~\cite{hoteit2014estimating} assumes that moving objects obey the uniform linear motion and use linear interpolation to obtain GPS points. DHTR~\cite{wang2019deep} recovers the GPS points of trajectory via a Seq2seq framework, followed by Kalman Filtering to reduce recovery uncertainty. 
Some works propose grid-based techniques~\cite{wei2012constructing, chen2011discovering} to divide the area into grids and find the shortest or the most popular path within the grid, followed by uniform sampling to recover GPS locations.
Subsequently, a map-matching algorithm is applied to project the recovery point onto the road network.}
\textcolor{black}{\textit{First map matching then recovering.} This solution first implements a map-matching algorithm on the sparse trajectory to project the GPS points onto the road network, representative methods like HMM-based methods~\cite{jagadeesh2017online, song2012quick} and learning-based methods~\cite{zhao2019deepmm, lou2009map, rappos2018force}. 
Subsequently, the shortest path on the road network is introduced in the second stage to obtain the complete trajectories.}
\textcolor{black}{However, both of the above two-stage solutions suffer from error accumulation, since both recovering GPS points and performing map-matching on the original sparse trajectories is unreliable.
Meanwhile, they are inefficient as the necessary map-matching algorithm in two-stage solutions is time-consuming.}



Another more efficient solution that has been recently proposed is the end-to-end method. This approach addresses the map-constrained trajectory recovery as a multi-tasking problem by aiming to directly recover both the sequence of road segments and the corresponding moving ratios. MTrajRec~\cite{ren2021mtrajrec} introduces a seq2seq multi-task pipeline that utilizes GRU to capture the temporal dependence of the trajectory data. RNTrajRec~\cite{chen2023rntrajrec} takes into account the correlation between trajectory and road network and uses the transformer to capture the dependence of trajectory points for trajectory recovery.
However, these methods adopt sequential models to encode trajectories, which fail to fully consider the trajectory's micro-semantics.

\textcolor{black}{A similarity task with map-constrained trajectory recovery is path-level trajectory recovery~\cite{liu2023graphmm, si2023trajbert, lin2021vehicle, yu2022spatio}, which only reconstructs the missing road segments from origin to destination.
Compared with them, map-constrained trajectory recovery simultaneously recovers the precise GPS location and road-related information, which provides more comprehensive information about trajectories, making it more advantageous for analyzing user movement patterns, understanding urban dynamics, and supporting various trajectory-related tasks.}

\subsection{Trajectory Modeling Technology}
\textcolor{black}{Trajectory data is commonly recorded in a sequential format, and many methods utilize sequential models to model trajectories. For instance, T2vec~\cite{li2018deep} utilizes Bi-LSTM~\cite{hochreiter1997long} to capture temporal correlations among historical and future GPS points, treating the final step's hidden state as the trajectory embedding. T3s~\cite{yang2021t3s} combines self-attention with LSTM~\cite{hochreiter1997long} to model global temporal correlations. NeuTraj~\cite{yao2019computing} integrates a spatial memory network into an RNN unit to capture spatial-temporal correlations within trajectories. Traj2vec~\cite{yao2017trajectory} utilizes a feature sequence to represent GPS trajectories and trains a sequence-to-sequence model to obtain low-dimensional vectors. AttnMove~\cite{xia2021attnmove} leverages attention mechanisms and human historical trajectories for predicting the next point of interest. START~\cite{jiang2022self} incorporates the self-supervised learning paradigm and attention mechanisms into trajectory representation learning.} \textcolor{black}{In summary, all the above sequence-based approaches only focus on capturing the explicit absolute information of a single GPS point. }

Although some methods like TrajGAT~\cite{yao2022trajgat} and PeriodicMove~\cite{sun2021periodicmove} model trajectory from a graph perspective, they focus on capturing spatial-temporal correlation of GPS points while ignoring the implicit relative information between GPS points, which limits their ability to fully encode the micro-semantics inherent within an individual trajectory.

\section{Preliminaries}
In this section, we first present several basic concepts and definitions related to the map-constrained trajectory task. Then, the frequently used notations in this paper are summarized in Table~\ref{tab:notation} to provide convenience for quick reference by the readers.

\subsection{Basic Definition}
\begin{definition}
[Trajectory] 
We define a trajectory as a sequence of GPS with timestamps, i.e., $\tau=\langle p_1, p_2, \cdots, p_N \rangle$, where each trajectory point $p_i=(lat_i, lng_i,t_i)$, $1\le i\le N$, represents that the GPS position of the moving object is $(lat_i, lng_i)$ at timestamp $t_i$. 
The sampling interval of $\tau$ is $t_i - t_{i - 1}, i \in \{2,\cdots, N\}$.
\end{definition}
\begin{definition}
[Road Network] 
A road network is defined as a directed graph $\mathcal{G}_r = (\mathcal{V}_r, \mathcal{E}_r)$, where $\mathcal{V}_r$ is the set of nodes representing intersections between road segments and each node has $lat$ and $lng$ attributes. 
$\mathcal{E}_r$ is the set of edges representing road segments that connect two nodes. An edge $e \in \mathcal{E}_r$ is decided by a start intersection $e.start \in \mathcal{V}_r$ and an end intersection $e.end \in \mathcal{V}_r$. 
\end{definition}
\begin{table}[t]	
	\centering
	\caption{Summary of frequently used notations.}
 \resizebox{1.0\columnwidth}{!}{
	\begin{tabular}{c|c}
		\toprule
		\textbf{Notations} & \textbf{Definition} \\
		\hline
		$p$ &  Sampled GPS point.\\
		$q$ & Map-constrained trajectory point.\\
		$e$ & Road segment.\\
		$r$ & Moving ratio.\\
		\hline
		$\tau$ & The observed sparse trajectory.\\
		$\tilde{\tau}$ & The recovered map-constrained trajectory.\\
            \textcolor{black}{$\mu$, $\epsilon$ }& 
        \textcolor{black}{The sampling interval of trajectory $\tau$ and $\tilde{\tau}$.}\\
		\hline
		$\mathcal{G}_r$ & The topology structure of the road network.\\
		$\mathcal{G}_{\tau}$ & The micro spatial-temporal trajectory graph.\\
		$\mathcal{G}_{f}$ & The macro trajectory flow graph.\\
		\hline
		\multirow{2}{*}{$\hat{e}^k_t$} & The road segment with the $k$-th highest probability\\
            & at time step $t$.\\
            \hline
		\multirow{2}{*}{$\mathcal{N}_{\hat{e}_{t}^k} $}  & The neighbors of the road segment with the $k$-th highest\\
        & probability at time step $t$ in the macro trajectory flow graph.\\
		\hline
		\multirow{2}{*}{$\hat{\textbf{h}}_i$} & The representation for the $i^{th} $ node\\
        & in the micro spatial-temporal trajectory graph.\\
        \hline
	\multirow{2}{*}{$\textbf{z}_{i,j} $} & The representation for the edge connecting  $i^{th} $ and $j^{th} $ \\
        & nodes in the micro spatial-temporal trajectory graph.\\
        \hline
	\multirow{2}{*}{$\textbf{M}_e$} & The representation for the road segment $e$ \\
        & in the macro trajectory flow graph.\\
        \hline
		$\textbf{RC}$ & The road condition representation.\\
		\bottomrule
	\end{tabular}
 }
	\label{tab:notation}
\end{table}

\begin{definition}
[Map-constrained Trajectory Point]
Due to sensor errors, the raw trajectory points are not necessarily strictly constrained on the road network. 
To address this, we define the map-constrained point of original trajectory point $p$ as $q=(e,r,t)$, where $q.t=p.t$ and $e \in \mathcal{E}_r$ represents the road segments where the map-constrained point is located, and $r \in [0,1]$ is the moving ratio which represents the ratio of moving distance along the road segment relative to its total length,
e.g., $r=0$ represents the point $q$ is located at the start point of the road segment $e$, $r=1$ indicates the point $q$ is located at the end of the road segment $e$.
Figure~\ref{fig:example} further illustrates the moving ratio. 
According to the road segment $e$ and moving ratio $r$, we can calculate the latitude and longitude of the map-constrained trajectory point $q$ as follows:
\begin{equation}
\label{eq:lat}
\begin{split}
    q.lat &= e.start.lat + (e.end.lat - e.start.lat) \ast r,\\
    q.lng &= e.start.lng + (e.end.lng - e.start.lng) \ast r.
\end{split}
\end{equation}

\end{definition}
\begin{definition}
[$\epsilon$-Sampling Interval Map-constrained Trajectory] A $\epsilon$-sampling interval map-constrained trajectory is a sequence of map-constrained trajectory points, defined as $\tilde{\tau }=\langle q_1,\cdots,q_m \rangle$. $\epsilon$ represents the time intervals between two successive map-constrained trajectory points, i.e., $\forall ~ i \in [2,m], q_i.t - q_{i-1}.t = \epsilon$.
We call $\tilde{\tau }$ as $\epsilon$-MM trajectory.
\end{definition}

\subsection{Problem Definition}
\textcolor{black}{Given a sparse trajectory $\tau$ with a sampling interval $\mu$ (e.g., the green GPS points $p_1, p_2, p_3$ in Figure~\ref{fig:example}) and an expected sampling interval $\epsilon$, our goal is to design an algorithm to recover the $\epsilon$-sampling interval map-constrained trajectory $\tilde{\tau }$. 
e.g., the red points $q_1 \sim q_7$ in Figure~\ref{fig:example}. 
Note, the sampling interval $\mu$ is larger than $\epsilon$.
In our study, we focus on recovering map-constrained trajectories with $\epsilon = 15$ seconds from the sparse trajectories with a sampling interval of 1 minute, 2 minutes, and 4 minutes respectively.}

\begin{figure*}
	\centering
	\includegraphics[width=0.99\textwidth]{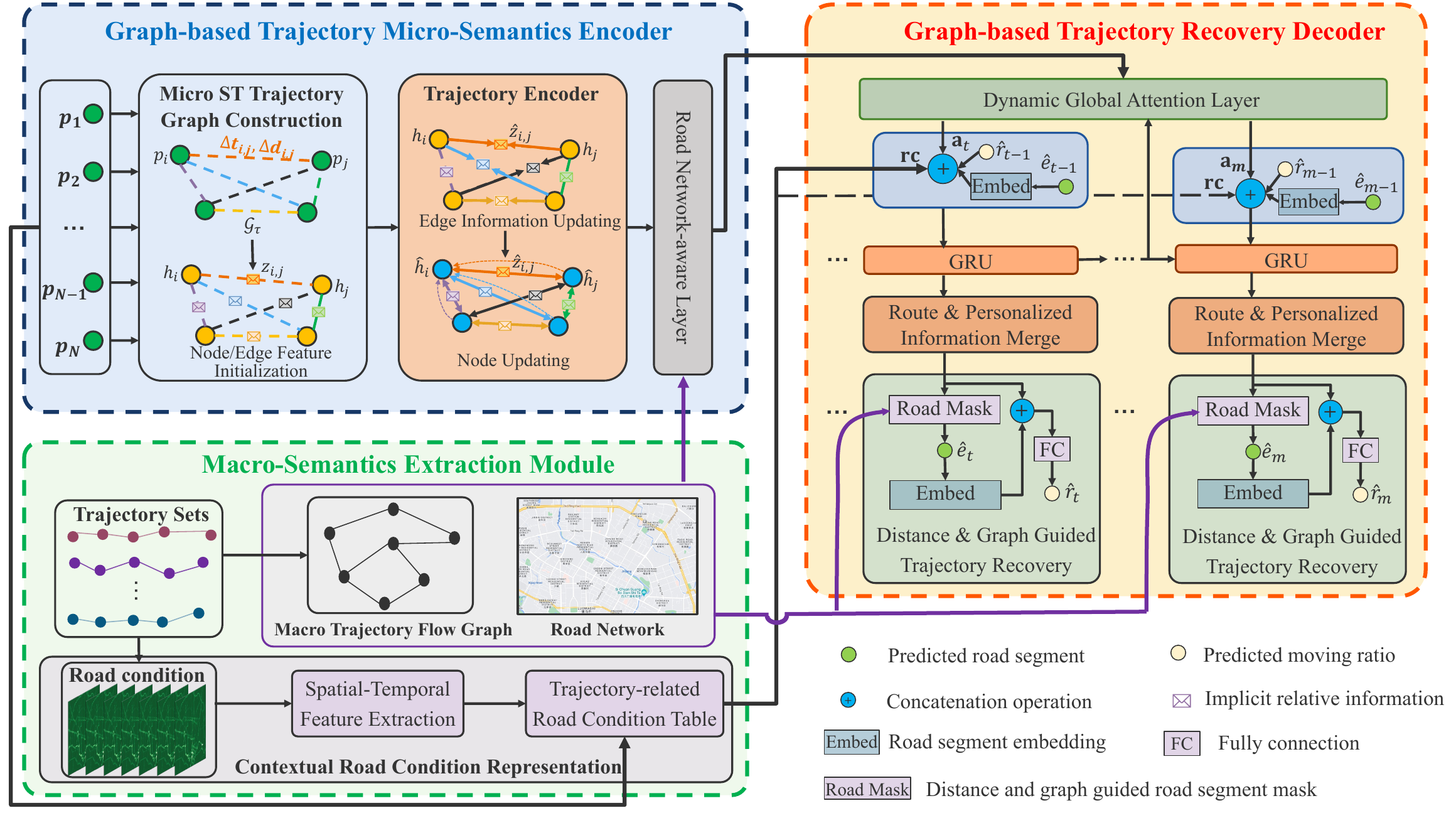}
	\caption{\textcolor{black}{The overall framework of MM-STGED. It consists of a graph-based trajectory micro-semantics encoder, a macro-semantics extraction module, and a graph-based trajectory recovery decoder.}
	}
	\label{fig:model}
\end{figure*}

\section{Methodology}
In this section, we introduce the proposed Micro-Macro Spatial-Temporal Graph-based Encoder-Decoder (MM-STGED) model, to tackle the map-constrained trajectory recovery task. We first provide an overview of the framework, and then describe each component in detail.

\subsection{Model Overview}
\textcolor{black}{Figure~\ref{fig:model} shows our proposed model. Given a sparse trajectory with a sampling interval $\mu$, our goal is to develop an end-to-end model that recovers the corresponding $\epsilon$-sampling interval map-constrained trajectory. The working pipeline of our model is described below.}

\textcolor{black}{First, we input a sparse trajectory into the graph-based trajectory micro-semantics encoder to extract its micro-semantics. In this module, we regard each trajectory as a graph, where the graph node is the GPS point that represents the explicit absolute information, and the edge represents the implicit relative information between these points. Then, we use a novel graph convolution mechanism to update the node and edge to obtain trajectory embedding and incorporate the surrounding road network to comprehensively capture the micro-semantic information embedded in trajectories.}

\textcolor{black}{At the same time, we input the overall trajectory data into the macro-semantics extraction module to derive macro-level information, including the human travel preferences and the traffic conditions related to the traveling routes. Specifically, based on the overall trajectories, we calculate the total number of consecutive passes between two road segments and construct a macro trajectory flow graph to represent human travel preferences. Meanwhile, we respectively divide the area of interest into grids in the spatial dimension and time into slices in the temporal dimension and calculate the flow within each cell. Finally, according to GPS points of the sparsity trajectory and the spatial-temporal index, we obtain trajectory-related road conditions.}

\textcolor{black}{Once we obtain the trajectory's micro- and macro-semantics, we input them together into the graph-based trajectory recovery decoder to obtain the expected recovered trajectory. In the graph-based trajectory recovery decoder, we employ an attention-based GRU as a fundamental component to recover trajectories in an autoregressive manner. 
At each decoding step, we take trajectory-related road conditions, recovered road segment ID $e_{t-1}$ and moving ratio $r_{t-1}$ of the previous as inputs, and generate the predictions of the current step's road segment ID $e_t$ and moving ratio $r_t$. 
In each step of the recovery process, to narrow down the search space of road segment candidates for recovery and improve the accuracy of the results, we divide GPS points into two categories: observed and unobserved GPS points. 
If the current step's GPS point is observed, we consider the surrounding road segments as candidates for recovery. Otherwise, we use the neighbor of $e_{t-1}$ in the macro trajectory flow graph as the candidates for recovery.}

\subsection{Graph-based Trajectory Micro-Semantics Encoder}
Different from previous trajectory recovery methods that encode trajectories using sequential models, we represent trajectories as graphs to capture micro-semantic information. 
Specifically, we construct a novel trajectory graph whose nodes correspond to trajectory points, and edges describe the implicit relative information between these points.
Then we design a graph-based trajectory encoder to embed trajectories by fully utilizing both the nodes and edges of the graph.

\subsubsection{Micro Spatial-Temporal Trajectory Graph Construction}
\textcolor{black}{Consider a sparse trajectory} $ \tau= \langle p_1, \cdots , p_N \rangle$, we represent it using a micro spatial-temporal trajectory graph $\mathcal{G_\tau} = (\mathcal{V}_\tau, \mathcal{E}_\tau, \textbf{A}^{\mathrm{time}}, \textbf{A}^{\mathrm{loc}})$, where each node corresponds to a trajectory point $p_i$ and $|\mathcal{V}_\tau| = N$.  
$\mathcal{E}_\tau =\{\mathcal{E}_{\tau }^{\text{time}} \cup  \mathcal{E}_{\tau }^{\text{loc}}\}$ is the edge set, $\mathcal{E}_{\tau }^{\text{time}}$ and $\mathcal{E}_{\tau }^{\text{loc}}$ are defined by $\textbf{A}^{\mathrm{time}}$ and $\textbf{A}^{\mathrm{loc}}$ respectively.
$\textbf{A}^{\mathrm{time}} \in \mathbb{R}^{N \times N}$ and $\textbf{A}^{\mathrm{loc}} \in \mathbb{R}^{N \times N}$ represent the time cost and space transfer adjacent matrix, the $(i, j)$-th element $a^{\mathrm{time}}_{i,j} \in \textbf{A}^{\mathrm{time}}$ and $a^{\mathrm{loc}}_{i,j} \in \textbf{A}^{\mathrm{loc}}$ are calculated as:
%
\begin{equation}
\begin{aligned}
\begin{split}
    a^{\mathrm{time}}_{i,j} = e^{-|t_i-t_j|},\quad a^{\mathrm{loc}}_{i,j} = e^{-\frac {dist_{i,j}} {\sigma} },
\end{split}
\end{aligned}
\end{equation}
where $dist_{i,j}$ is the distance between point $p_i$ and $p_j$, and $\sigma$ denotes the variance in distances across the trajectory $\tau$.
\textcolor{black}{Each trajectory produced by a traveler exhibits inherent characteristics associated with the purpose and pattern of travel, leading to the existence of spatial-temporal correlations among all the points within the trajectory. To comprehensively capture these correlations, our micro spatial-temporal trajectory graph $\mathcal{G}_\tau$ is fully connected.}


\noindent \textbf{Node/Edge feature initialization.}
The simplest option to initialize node features is to feed the latitude and longitude of each trajectory point into a Multi-Layer Perceptron (MLP), but this method has two shortcomings:
\textcolor{black}{(1) The differences along the latitude and longitude between consecutive points in the trajectory are often tiny, which may confuse the model in the prediction tasks~\cite{zheng2016generating}.
(2) The trajectory is inherently continuous in nature, which implies that these trajectory points are not independent entities but are intrinsically linked through spatial-temporal correlations. However, MLP regards the input entities as independent and does not consider their correlation.}


To overcome these drawbacks, we first partition the area of interest into grids and represent each trajectory point $p_i$ as a triplet $(x_i,y_i,tid_i)$, where $x_i$ and $y_i$ denote the indices of the grid cell in which $p_i$ is located. $tid_i=\left \lfloor \frac{t_i-t_1}{\epsilon}  \right \rfloor $ is the time index of trajectory point $p_i$ under the desired sampling interval $\epsilon$. 
Introducing $tid_i$ makes the model aware of the number of points that are to be reconstructed between two consecutive observed trajectory points. 
Then, we use a gated recurrent unit (GRU) ~\cite{cho2014learning} to model the temporal correlations between discrete trajectory points.
During the processing of each trajectory point $p_i$, the GRU receives the grid cell indices $x_i, y_i$, the time index $tid_i$, and the hidden state $\textbf{h}_{i-1}$ from the previous step as input, and yields the hidden state $\textbf{h}_i$ as the initialized feature for node $i$. 


The edges in graph $\mathcal{G}_\tau $ reveal the underlying implicit relative information between trajectory points. Specifically, each edge is characterized by two raw attributes, that is,  
$\textbf{e}_{i,j}=a^{\mathrm{time}}_{i,j} \parallel 
 a^{\mathrm{loc}}_{i,j}$,
where $\parallel$ is the concatenation operator.
We employ a non-linear transformation to project the raw edge feature into $d$-dimension: 
\begin{equation}
    \textbf{z}_{i,j} = \sigma (\textbf{W}_e \textbf{e}_{i,j} + \textbf{b}_e ),
\end{equation}
where $\sigma$ is ReLU activation function, $\textbf{W}_e \in \mathbb{R}^{d\times 2}$ and $\textbf{b}_e \in \mathbb{R}^{d}$ are learnable parameters.

\subsubsection{Trajectory Encoder}
Graph convolution networks (GCNs)~\cite{welling2016semi} are widely used to encode graph structure data. Traditional GCNs perform message-passing only among the node
features, guided by the topological adjacency matrix. Such GCNs do not take into account the rich edge features.
However, in our case, 
\textcolor{black}{both node features and edge features}
play a pivotal role in describing the micro-semantic information of individual trajectory, which is valuable for the trajectory recovery task.    
Therefore, we carefully design a novel graph convolution mechanism that simultaneously fuses and updates nodes and edges information. 

Given the micro spatial-temporal trajectory graph $\mathcal{G}_\tau$, we jointly encode the node and edge embeddings to fully describe the micro-semantic information. Specifically, the embedding of an edge, denoted as $\textbf{z}_{i,j}$, is affected by itself and those nodes it connects. Meanwhile, the embedding of a node is affected by its own attributes, its neighbors, as well as the edges interconnecting them. Formally: 
\begin{equation}
    \hat{\textbf{z}}_{i,j} = \textbf{W}_1\textbf{z}_{i,j} + \textbf{W}_2\textbf{h}_i + \textbf{W}_3\textbf{h}_j,
\end{equation}
\begin{equation}
    \hat{\textbf{h}}_{i} = f(\textbf{h}_i, \text{Agg}(\textbf{h}_j,\hat{\textbf{z}}_{i,j}|j\in\mathcal{N}_i) ),
\end{equation}
where $\mathcal{N}_i$ denotes the neighbor set of node $i$. $\text{Agg}(\cdot)$ is the aggregation function and $f(\cdot)$ is the updating function, which we implement as follows: 
\begin{equation}
\small
    \hat{\textbf{h}}_i = \textbf{h}_i + \sigma (\text{BN}(\textbf{W}_4\textbf{h}_i + \textbf{W}_5\hat{\textbf{h}} ^{\mathrm{time}}_i +\textbf{h}_6 \hat{\textbf{h}}^{\mathrm{loc}}_i + \textbf{W}_7 \sum_{j\in\mathcal{N}_i }\hat{\textbf{z}}_{i,j})),
\end{equation}
where 
$\text{BN}(\cdot)$ denotes batch normalization. 
Here we use the residual connection to pass through the raw node features. $\hat{\textbf{h}}_i^{\mathrm{time}}$ and $\hat{\textbf{h}}_i^{\mathrm{loc}}$ are aggregated temporal neighbor and spatial neighbor features, respectively, which are calculated as:
\begin{equation}
    \hat{\textbf{h}}_i^{\mathrm{time}} = \sigma(\textbf{A}^{\mathrm{time}} \textbf{h}_i\textbf{W}_{8}), 
\end{equation}
\begin{equation}
    \hat{\textbf{h}}^{\mathrm{loc}}_i = \sigma(\textbf{A}^{\mathrm{loc}} \textbf{h}_i\textbf{W}_{9}),
\end{equation}
where $\textbf{W}_{\ast}\in \mathbb{R}^{d\times d}, \ast \in\{1,\cdots,9\}$ are learnable parameters. As graph $\mathcal{G}_\tau $ is fully connected, only performing the aggregation process once is able to make each node perceive the information from other nodes and their edges.


\subsubsection{Road Network-aware Layer}
\textcolor{black}{While the above process effectively incorporates longitude, latitude, and timestamp as explicit absolution information, and space transfer and time cost as implicit relative information in learning the representations for trajectory points, it does not model the surrounding road network, which is also an important aspect of the trajectory's micro-semantics.}
As discussed in Section~\ref{sec1}, the road network plays a critical role in affecting the user's route choice. Therefore, we propose a road network-aware layer to enhance trajectory representation.

Specifically, we first use a macro trajectory flow graph (see Section~\ref{flow_graph} for details) to represent the road network, since this representation is capable of comprehensively reflecting both the physical topology of the road network and human travel preferences. Then we use Node2vec~\cite{grover2016node2vec} to learn the representations of the road segments $\textbf{M} \in \mathbb{R}^{|\mathcal{R}| \times D}$, where $|\mathcal{R}|$ represents the number of road segments. 
For each of the observed trajectory points $p_i$, we define a distance weighting function $f(d_{i,l})$ to quantify the correlation between $p_i$ and a road segment $l \in \mathcal{R}$. The function $f(d_{i,l})$ is formulated as:
\begin{equation}\label{f_dis} 
	f(d_{i,l}) =   \begin{cases}
		\mathrm{exp}({-(d_{i,l})^2 / \kappa ^2}),  & \text{if }   d_{i,l} < \epsilon_{dist},\\
		0,  & \text{otherwise}.
	\end{cases}
\end{equation}
where $d_{i,l}$ is the shortest distance between point $p_i$ and road segment $l$, and $\kappa$ is a hyper-parameter. Obviously, the function $f(d_{i,l})$ takes into account only the road segments whose distance to $p_i$ is below the threshold $\epsilon_{dist}$, while the road segments are disregarded if their distances to $p_i$ exceed this threshold.
We then get the road network representation $\textbf{h}_i^{\text{road}}$ for each trajectory point $p_i$ by: 
\begin{equation}
	\textbf{h}_{i}^{\text{road}} = \sum_{l=1}^{|\mathcal{R}|} f(d_{i,l}) * \textbf{M}_l.
\end{equation}

Finally, we concatenate the representation for each trajectory point output from the trajectory encoder and that output from the road network-aware layer together, then employ a non-linear transformation to obtain the final representation $\bar{\textbf{h}}_i\in \mathbb{R}^d$ for each trajectory point,
\begin{equation}
    \bar{\textbf{h}}_i = \sigma (\textbf{W}_{h}[\hat{\textbf{h}}_i || \textbf{h}_{i}^{\text{road}}] + \textbf{b}_{h}),
\end{equation}
where $\textbf{W}_{h}$ and $\textbf{b}_{h}$ are learnable parameters. In addition, we use a mean pooling layer to read out the embedding of the whole trajectory.

\begin{figure}
	\centering
	\includegraphics[width=0.45\textwidth]{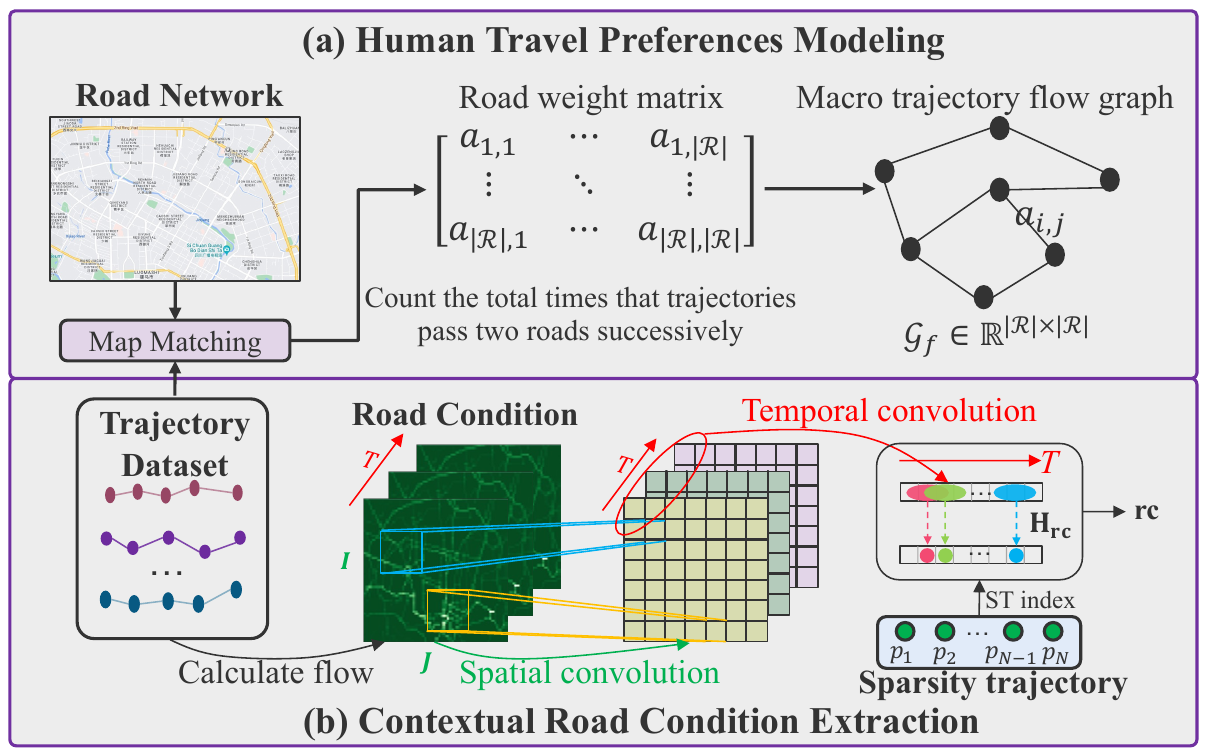}
	\caption{\textcolor{black}{Illustration of macro-semantics extraction.}
	}
	\label{fig:macro}
\end{figure}
\subsection{Macro-Semantics Extraction Module}
To effectively extract the macro-semantic information reflected by a group of trajectories, we propose a twofold scheme that incorporates: a macro trajectory flow graph that models human travel preferences and a road condition representation that describes the environment of the studied trajectory. Next, we introduce them in detail.

\subsubsection{Macro Trajectory Flow Graph Construction}
\label{flow_graph}
The macro trajectory flow graph is designed to provide a global view of people's travel preferences. 
\textcolor{black}{As shown in Figure~\ref{fig:macro}(a),} we first perform map matching for all observed trajectories based on the Hidden Markov Model (HMM)~\cite{newson2009hidden} to convert the trajectory point sequences into road segment sequences. Subsequent to this transformation, we construct the macro trajectory flow graph \textcolor{black}{based on the user's road segment selections.}

Formally, the macro trajectory flow graph is a weighted directed graph defined as $\mathcal{G}_f = (\mathcal{R}, \mathcal{E}, \textbf{A})$:
\begin{itemize}
	\item $\mathcal{R}$ is the node set. i.e., the set of road segments. 
	\item $\mathcal{E}$ is the edge set. There is an edge from $r_i$ to $r_j$ if there exists a trajectory that passes road $i$ and road $j$ successively. We add self-connection for each node.
	\item $\textbf{A}$ is the weight matrix. $a_{i,j} \in \textbf{A}$ equals the total times that trajectories pass road $i$ and road $j$ successively. 
\end{itemize}

\subsubsection{Contextual Road Condition Representation}
In real-world scenarios, the choice of travel routes and the corresponding travel patterns, such as speed, are inherently dynamic and are affected by actual contextual road conditions. 
For example, congestion on a chosen route may compel a driver to drive slower or even prompt them to seek alternative routes, both of which inevitably impact trajectory. 
In light of this, we propose to incorporate the influence of contextual road conditions into the proposed model, thereby simulating the environment of the trajectory studied.

To obtain road condition information, \textcolor{black}{as shown in Figure~\ref{fig:macro}(b),} we divide the area of interest into $I \times J$ grid cells in spatial dimension and partition the temporal scope into $T$ time steps. Through the analysis of historical trajectory records, we derive the traffic flow for each individual cell, represented by the matrix $\textbf{RC}\in \mathbb{R}^{I\times J\times T}$. 
To characterize the spatial-temporal correlations of road conditions, we employ a 1D convolution (1D CNN) along the temporal dimension and a 2D CNN in the spatial domain to process $\textbf{RC}$.
Finally, we get a comprehensive representation of road condition for all grid cells, denoted by $\textbf{H}_{\mathrm{rc}} \in \mathbb{R}^{I\times J \times T \times F}$, where $F$ is the dimension of the road condition embedding.

Considering that a trajectory is happening over a specific time period and covers a specific area, to avoid information redundancy, we selectively extract the related road condition information that is closely associated with the trajectory by employing the following aggregation formula: 
\begin{equation}
    \textbf{rc} = \frac{1}{N}\sum_{i=1}^{N} \textbf{H}_{\mathrm{rc}}[\pi_{\mathrm{lat}}(p_i.lat),\pi_{\mathrm{lon}}(p_i.lng),\pi_{t}(p_i.t)],
\end{equation}
where $\pi_{\mathrm{lat}}(\cdot ),\pi_{\mathrm{lng}}(\cdot ),\pi_{t}(\cdot )$ denote index functions that map the latitude, longitude, and time to their corresponding indices.
Later, the trajectory-related road condition \textbf{rc} is leveraged in the trajectory recovery process, thereby enabling the model to capture the dynamic nature of trajectories. 

\subsection{Graph-based Trajectory Recovery Decoder}
\textcolor{black}{
After the encoder has sufficiently processed the trajectory, the decoder recovers the $\epsilon$-MM trajectory in an auto-regressive manner. For this purpose, GRU is chosen as a foundational component. We explain the design of the decoder in detail as follows.
}

At each time step $t$, the input to the GRU consists of two parts: the hidden state of the previous step $\textbf{s}_{t-1}$, and the input of the current step. 
Specifically, the input of the current step includes the embedding of road segment $\textbf{x}_{e_{t-1}}$ and the moving ratio $r_{t-1}$ output from the previous step, the trajectory-related contextual road condition representation $\textbf{rc}$, and the weighted trajectory representation output from the encoder. 
Due to different points in the sparse trajectory contributing variably to the recovery of different $\epsilon$-MM trajectory points, we introduce a global attention mechanism that weights the contribution of each trajectory point based on its relevance to the current decoding step, formulated as: 
\begin{equation}
        \textbf{s}_t = \text{GRU}(\textbf{s}_{t-1}, \textbf{x}_{e_{t-1}}\parallel r_{t-1} \parallel \textbf{rc} \parallel \textbf{a}_t),
\end{equation}
\begin{equation}
        \textbf{a}_t = { \sum_{i=1}^{N}} a_{t,i}\hat{\textbf{h}}_i,
\end{equation}
\begin{equation}
        a_{t,i} = \frac{e^{\textbf{v}^\top \mathrm{tanh}(\textbf{W}_s\textbf{s}_{t-1}+\textbf{W}_h\hat{\textbf{h}}_i) }}
{\sum_{i'=1}^{N} e^{\textbf{v}^\top \mathrm{tanh}(\textbf{W}_s\textbf{s}_{t-1}+\textbf{W}_h\hat{\textbf{h}}_{i'})}}, 
\end{equation}
where $\textbf{v} \in \mathbb{R}^d,\textbf{W}_s,\textbf{W}_h \in \mathbb{R}^{d \times d}$ are learnable parameters. 
The first hidden state $\textbf{s}_{0}$ is initialized by the pooled trajectory embedding generated by the encoder.
Once the GRU produces the decoded hidden state $\textbf{s}_t$ for time step $t$, we can predict the road segment and the moving ratio of the corresponding map-constrained trajectory point based on $\textbf{s}_t$. 

A straightforward way to predict the road segment is to treat all the road segments as candidates and then perform multi-classification. However, this way will waste unnecessary computations and increase the uncertainty of finding the correct road segments, since it is impossible for many road segments to be the correct answer. 
In fact, trajectories are generated by people’s traveling. From a macro-perspective, people have specific travel patterns, a kind of prior knowledge, and the generated trajectories should also obey these specific patterns. Therefore, we could utilize macro-semantic information to narrow down the road segment candidates. 

Specifically, for a recovered trajectory point at time step $t$ under the sampling interval $\epsilon$,
there are two cases regarding its relationship with the raw trajectory $\tau$: 
\begin{itemize}
    \item Case 1: the GPS location at time step $t$ is observed. i.e. $\exists~{i}\in\{1,\cdots,N\}, p_i.tid=t$, where $p_i$ is the $i$-th sampled points in the raw sparse trajectory $\tau$. 
    \item Case 2: the GPS location at time step $t$ is not observed. i.e. $\forall~{i}\in\{1,\cdots,N\}, p_i.tid \ne t$. 
\end{itemize}

For case 1, theoretically, $p_i$ should be located on a certain road $l$. But in reality, $p_i$ may have a slight distance from road $l$ due to GPS device error, as shown in Figure~\ref{fig:example}. 
Thus, we use distance to select the road segment candidates, by which we can reduce the search space and improve our model’s efficiency and accuracy.
Here, we use the weighting function $f(d_{i,l})$ defined in Equation~\ref{f_dis} to represent the correlation between $p_i$ and road segment $l$ and select the road segments that are not far from $p_i$ as candidates. 

For case 2, we do not know the GPS location of the expected recovered trajectory point at time step $t$, so we cannot directly calculate its distance to each road segment.
As an alternative, we turn to resort to the predicted road segment $\hat{e}_{t-1}$ at the previous time step and the macro trajectory flow graph $\mathcal{G}_f$ to guide trajectory recovery. i.e., when making the prediction at time step $t$, we choose the neighbors of node $\hat{e}_{t-1}$ in $\mathcal{G}_f$ as candidates, since people’s travel habits tend to follow the global transfer preferences.

However, such auto-regressive decoders are prone to error accumulation. If $\hat{e}_{t-1}$ is wrong, the following recovered trajectory points will be in the wrong direction. 
To alleviate this
shortcoming, we consider a broader set of candidate road segments based on the top-$K$ highest probabilities from the previous time step, i.e., $\{\hat{e}_{t-1}^1, \hat{e}_{t-1}^2,\cdots,\hat{e}_{t-1}^K\}$. We gather the neighbors of all these road segments, denoted as $\{\mathcal{N}_{\hat{e}_{t-1}^1},\mathcal{N}_{\hat{e}_{t-1}^2}, \cdots, \mathcal{N}_{\hat{e}_{t-1}^K}\}$ as the road segment candidates at time step $t$, where $\hat{e}^k_{t-1}$ is the road segment with the $k$-th highest probability at time step $t-1$.
In this way, the road segment candidate space at time step $t$ will be expanded from $\mathcal{N}_{\hat{e}_{t-1}}$ to $\mathbb{S}_{t-1} = {\mathcal{N}_{\hat{e}_{t-1}^1} \cup \mathcal{N}_{\hat{e}_{t-1}^2} \cup \cdots \cup \mathcal{N}_{\hat{e}_{t-1}^K}}$, which covers all reasonable road segments, avoiding the over-reliance on the correctness of the predicted road segment in the last step.

In conclusion, we define the correlations between the expected recovered trajectory point at time step $t$ and a road segment $l$ as follows:
\begin{small}
\begin{equation}
\psi_{t,l} =   \begin{cases}
f(d_{i,l}),  & \text{if }   \exists ~ p_i \in \tau, p_i.tid = t \text{ and } d_{i,l} < \epsilon_{dist}, \\
1,  & \text{if } \forall ~ p_i \in \tau, p_i.tid \neq t \text{ and } l \in \mathbb{S}_{t - 1}, \\
0,  & \text{otherwise}.
\end{cases}
\end{equation}
\end{small} 

Then, we utilize this correlation to guide the prediction of road segments at time step $t$. 
Formally:
\begin{equation}
    \label{equ_p}
    \mathbb{P}(l|\textbf{s}_t, \mathcal{G}_f) = \frac{\mathrm{exp}(\textbf{s}_t^\top \textbf{w}_{l}) \ast \psi_{t,l} }
{\sum_{l'}^{} \mathrm{exp}({\textbf{s}_t^\top \textbf{w}_{l'}  }) \ast \psi_{t,l'}} ,
\end{equation}
where $\textbf{w}_l$ is the $l$-th column vector from a trainable parameter $\textbf{W}^{{l}} \in \mathbb{R}^{d \times |\mathcal{R}|}$. After obtaining the probability with each road segment, we employ the $\arg\max$ function to get the predicted road segment $\hat{e}_t$ at time step $t$. 
Subsequently, we feed the GRU hidden state $\textbf{s}_t$ and the embedding of $\hat{e}_t$ into a two-layer MLP equipped with a Sigmoid activation function, to get the moving ratio $\hat{r}_t$.

\noindent \textbf{Route and Personalized Information Merge.}
In practical scenarios, users' personalized preferences, as well as the road segments traversed before time step $t$ are helpful for trajectory recovery at time step $t$. 
Therefore, at each step, we utilize such information to enrich the hidden state $\textbf{s}_t$ for the decoder. 
    Specifically, taking into account the topological relations between road segments, we first apply Node2Vec~\cite{grover2016node2vec} on the macro trajectory flow graph $\mathcal{G}_f$ to obtain the embedding $\textbf{M}$ for each road segment. 
Then, we concatenate the user's embedding $\textbf{u}$ (randomly initialized and learnable vector), current step hidden state $\textbf{s}_t$, and the embeddings of previously traversed road segments. The concatenated representation, denoted as $\hat{\textbf{s}}_t$, to replace $\textbf{s}_t$ in Equation~\ref{equ_p}. Formally: 
\begin{equation}
    \hat{\textbf{s}}_t = \textbf{s}_t \parallel \textbf{u} \parallel \sum_{j=1}^{t-1}\textbf{M}_{\hat{e}_j}.
\end{equation}

\subsection{Model Training}
We train the model by simultaneously optimizing the road segment prediction and moving ratio prediction tasks. 
Cross-entropy is employed as the loss function for the road segment prediction:
\begin{equation}
    \mathcal{L}_1 = -  \frac{1}{|\tilde{\tau}|}\sum_{t=1}^{|\tilde{\tau}|}  \text{log  }\mathbb{P} (\hat{e}_t| \hat{\textbf{s}}_t,\mathcal{G}_f),
\end{equation}
where $|\tilde{\tau}|$ is the length of $\epsilon$-MM trajectory.
For the moving ratio prediction, we adopt the mean squared error as the loss function:
\begin{equation}
    \mathcal{L}_2 = \frac{1}{|\tilde{\tau}|}\sum_{t=1}^{|\tilde{\tau}|}   |\hat{r}_t - r_t|^2.
\end{equation}

In the end, our final loss function is $\mathcal{L} = \mathcal{L}_1 + \lambda \mathcal{L}_2$, 
where $\lambda$ is a hyper-parameter to balance the above loss.

\section{Experiments}
To evaluate the effectiveness of our proposed model, we conduct an extensive set of experiments covering three sampling intervals of sparse trajectory across two datasets. 

\subsection{Datasets}
We evaluate our model on two real-world trajectory datasets collected from Porto, Portugal, and Chengdu, China. 
Additionally, the road network is downloaded from OpenStreetMap\footnote{\url{http://www.openstreetmap.org/}}. 

The Porto dataset\footnote{\url{https://www.kaggle.com/competitions/pkdd-15-predict-taxi-service-trajectory-i/data}} contains trajectory data collected from July 2013 to June 2014 by 442 taxis with the sampling interval of 15 seconds.
The Chengdu dataset\footnote{\url{https://outreach.didichuxing.com/}} records trajectory data collected during November 2016 by 17675 taxis, and has an initial sampling interval of 2 to 3 seconds. Here, we set the sampling interval to around 15 seconds by resample. 
The road network retrieved from OpenStreetMap is highly detailed, however, some roads (such as footways) are inaccessible to vehicles. Therefore, we used ArcGIS software to filter out these inaccessible road types by visualizing the road network and trajectory points For the Porto dataset, we retain four road types: ``Primary, Secondary, Motorway, Tertiary". For the Chengdu dataset, we opt for ``Primary, Secondary, Trunk, Tertiary". After selecting the road types, we employ a map-matching algorithm~\cite{newson2009hidden} on the raw trajectories to get the ground truth. The detailed statistical information of datasets is summarized in Table~\ref{tab:dataset}.

\begin{table}[h]	
\small
\centering
\caption{Dataset Description.}
\begin{tabular}{c|c|c}
    \toprule
    Types & Porto & Chengdu \\
    \hline
    Sampling interval & 15s & 15s \\
    \# Trajectory & 322,079 & 118,354 \\
    \# Road segment & 2224 & 2504 \\
    Latitude range & (41.142, 41.174) & (30.655, 30.727) \\
    Longitude range & (-8.652, -8.578) & (104.043, 104.129) \\
    Length / Width & 6.19 km / 3.56 km & 8.22 km / 8.00 km\\
    \# User & 442 & 17675 \\
    \bottomrule
\end{tabular}
\label{tab:dataset}
\end{table}

\subsection{Evaluation Metrics}
We evaluate the performance of MM-STGED by using five standard metrics. Out of which three - \textbf{Accuracy (Acc)}, \textbf{Recall}, and \textbf{Precision (Prec)} - are used to evaluate the model's accuracy in predicting road segments by comparing the recovered road segments $\mathcal{E}_{\hat{p}} = \{\hat{e}_1, \hat{e}_2, \cdots, \hat{e}_m \}$ and the ground-truths $\mathcal{E}_{\hat{p}} = \{e_1, e_2, \cdots, e_m \}$. A higher value for these three metrics signifies better accuracy in road segment prediction. These metrics are formulated as: 
\begin{equation}
\begin{split}
	\textbf{Acc} &= \frac{1 }{m}  {\sum_{i=1}^{m}} \mathbbm{1}(e_i, \hat{e}_i) \times 100 \%,\\
 \textbf{Recall} &= \frac{|\mathcal{E}_{\hat{p}}\cap\mathcal{E}_p | }{|\mathcal{E}_p|}\times 100 \%, \\
 \textbf{Prec} &=  \frac{|\mathcal{E}_{\hat{p}}\cap\mathcal{E}_p | }{|\mathcal{E}_{\hat{p}}|}\times 100 \%,
\end{split}
\end{equation}
where $\mathbbm{1}(\cdot)$ is the indicator function, and $\mathbbm{1}(e_i, \hat{e}_i)$ equal 1 if $e_i=\hat{e}_i$ else 0.

To evaluate the accuracy of the recovered GPS location, we use \textbf{Mean Absolute Error (MAE)} and \textbf{Root Mean Square Error (RMSE)} to quantify the distance between the recovered trajectory point $\hat{q}$ and the ground-truth point $q$. It should be noted that $\hat{q}$ can be calculated from the predicted road segment and moving rate by Equation~\ref{eq:lat}.
Following MTrajRec~\cite{ren2021mtrajrec}, the errors are computed as the shortest distance on the road network. The \textbf{MAE} and \textbf{RMSE} are calculated as:
\begin{equation}
\begin{split}
    \textbf{MAE} &= \frac{1}{m}\sum_{j=1}^{m}|\mathrm
        {dist}(\hat{q}_j,q_j )|, \\
    \textbf{RMSE} &= \sqrt{\frac{1}{m}\sum_{j=1}^{m}|\mathrm{dist}(\hat{q}_j,q_j )|^2  },
\end{split}
\end{equation}
where $m$ is the number of points in $\epsilon$-MM trajectory, $\mathrm{dist}(\hat{q}_j, q_j)$ represents the shortest road network distance between $\hat{q}_j$ and $q_j$. Lower values of $\textbf{MAE}$ and $\textbf{RMSE}$ imply superior model performance.
Both \textbf{MAE} and \textbf{RMSE} are expressed in meters.

\subsection{Baselines}
To prove the superiority of the proposed method, we establish the following baselines: 
\begin{itemize}
     \item \textcolor{black}{\textbf{HMM~\cite{newson2009hidden} + ShortestPath} first employs the map-matching algothrim on the sparse trajectory, then utilizes the shortest path to determine the missing trajectory points.}
	\item \textbf{Linear~\cite{hoteit2014estimating} + HMM~\cite{newson2009hidden}} first uses linear interpolation to get the recovered trajectory and then introduces HMM to project trajectory onto the road network.
 \item \textcolor{black}{\textbf{MPR~\cite{chen2011discovering} + HMM~\cite{newson2009hidden}} first grids the map to identify the most popular route among the grids and interpolates the trajectory points assuming that the vehicle moves at a constant speed, then implements the HMM to project these points onto the road network.}
	\item \textbf{DHTR~\cite{wang2019deep} + HMM~\cite{newson2009hidden}} first recovers trajectory based on the seq2seq model and Kalman filter, and then implements HMM to perform the map matching process.
	\item \textbf{AttnMove~\cite{xia2021attnmove}} \textbf{+ Rule} utilizes the attention module to predict the recovery road segments and uses the central location as the final result. 
	\item \textbf{MTrajRec~\cite{ren2021mtrajrec}} is an end-to-end method to do trajectory recovery task, which first utilizes GRU to encoder trajectory, and then proposes multi-task learning recovering trajectory points autoregressively.
	\item \textbf{T2vec~\cite{li2018deep} + Decoder~\cite{ren2021mtrajrec}} is a deep learning method for trajectory similarity learning, here we choose its encoder to learn trajectory embedding, and then we use the decoder proposed in MTrajRec to get the results.
	\item \textbf{T3s~\cite{yang2021t3s} + Decoder~\cite{ren2021mtrajrec}} capture the spatial-temporal correlation of the trajectory by an LSTM and self-attention network, and then combine the decoder proposed in MTrajRec to recover the trajectory.
	\item \textbf{RNTrajRec~\cite{chen2023rntrajrec}} is the state-of-the-art method for trajectory recovery, which uses a transformer and road network to capture the spatial-temporal correlation between trajectory points.
\end{itemize}


\subsection{Experimental Settings}
\textcolor{black}{We split the datasets into training sets, validation sets, and testing sets at a ratio of 7:2:1. 
We employ the HMM algorithm~\cite{newson2009hidden} to project the original GPS points to map-constrained trajectory points, yielding map-constrained trajectories with a sampling interval of 15 seconds. 
To obtain the sparse trajectory, we resample the original trajectory by setting the time intervals of 4 minutes, 2 minutes, and 1 minute to construct three types of sparse trajectories.
This equals that we use 6.25\%, 12.5\%, and 25\% trajectory points to recover the rest.}


We implement MM-STGED\footnote{\url{https://github.com/wtl52656/MM-STGED}} using PyTorch \cite{paszke2019pytorch} and train the model for 20 epochs with a batch size of 128 and a learning rate of 1e-3. We set the model dimension $d = 512$, and the embedding dimension for Node2Vec $D = 64$. To take road conditions into account, we divide the map into a 64 $\times$ 64 grid and set the grid dimension to $F = 64$. In Equation~\ref{f_dis}, we set $\kappa = 15$, $\epsilon_{dist} = 50$ meters. In the loss function, we set $\lambda = 10$ to balance the two tasks.
\begin{table*}[h]	
\centering
\caption{Performance comparison on two datasets with sampling intervals at 4 minutes, 2 minutes, and 1 minute, respectively.}
\resizebox{1.0\linewidth}{!}{
\begin{tabular}{c|c|ccccc|ccccc}
    \toprule
    \multirow{2}{*}{Sampling Interval}&\multirow{2}{*}{Methods}&\multicolumn{5}{c|}{Chengdu} &\multicolumn{5}{c}{Porto}\\
    \cline{3-12}
    & & Acc($\%$) & Recall($\%$) & Prec($\%$) & MAE & RMSE & Acc($\%$) & Recall($\%$) & Prec($\%$) & MAE & RMSE\\
    \midrule
    \multirow{10}{*}{\parbox{2.5cm}{\centering $\mu = 4$ minutes $\to \epsilon = 15$ seconds}} & HMM + ShortestPath & 26.85 & 28.64 & 29.55 & 939.3 & 1047.7 & 20.19 & 26.22 & 33.51 & 886.9 & 941.5 \\
    & Linear + HMM & 26.42 & 30.45 & 36.15 & 974.5 & 1145.4 & 32.23 & 36.09 & 49.80 & 489.3 & 637.3\\
    & MPR + HMM & 36.93 & 38.62 & 44.53 & 821.9 & 914.1 & 32.22 & 38.67 & 48.07 & 534.0 & 700.2\\
    & DHTR + HMM & 41.48 & 57.34 & 50.48 & 673.6 & 911.3 & 32.02 & 58.29 & 45.61 & 456.7 & 627.1 \\
    & AttnMove + Rule & 63.43 & 73.97 & 78.72 & 358.2 & 916.7 & 49.31 & 48.62 & 78.03 & 310.0 & 621.3\\
    & MTrajRec & 65.79 & 75.14 & 78.42 & 315.1 & 904.4 & 52.36 & \underline{60.39} & 77.28 & 266.1 & 590.1\\
    & T3s + Decoder & 65.60 & 75.26 & 78.14 & 318.2 & 926.3 & 52.24 & 60.24 & 77.80 & 270.4 & 594.9 \\
    & T2vec + Decoder & 66.51 & \underline{75.68} & 78.27 & 307.5 & 915.2 & 53.13 & 60.27 & 77.62 & 256.3 & 571.0\\
    & RNTrajRec & \underline{67.66} & 75.59 & \underline{79.97} & \underline{306.1} & \underline{886.0} & \underline{54.59} & \textbf{60.42} & \underline{79.20} & \underline{248.1} & \underline{549.1}\\
    & \textbf{MM-STGED (Ours)} & \textbf{70.64} & \textbf{76.04} & \textbf{81.63} & \textbf{266.2} & \textbf{829.7} & \textbf{57.30} & 59.48 & \textbf{80.21} & \textbf{222.8} & \textbf{510.4}\\
    \midrule
    \multirow{10}{*}{\parbox{2.5cm}{\centering $\mu = 2$ minutes $\to \epsilon = 15$ seconds}} & HMM + ShortestPath & 33.85 & 47.89 & 48.31 & 754.1 & 826.2 & 27.30 & 40.05 & 46.00 & 647.3 & 747.0\\
    & Linear + HMM & 43.78 & 45.35 & 48.77 & 816.9 & 1054.7 & 49.35 & 50.45 & 63.87 & 408.7 & 609.8\\
    & MPR + HMM & 49.88 & 54.62 & 50.94 & 474.8 & 899.0 & 49.76 & 52.97 & 62.86 & 409.8 & 610.9\\
    & DHTR + HMM & 47.17 & 60.16 & 51.73 & 662.0 & 912.2 & 43.76 & 65.25 & 52.10 & 385.3 & 578.0\\
    & AttnMove + Rule & 71.98 & 77.42 & 80.67 & 291.1 & 764.8 & 61.39 & 60.90 & 82.98 & 213.4 & 468.5\\
    & MTrajRec & 74.52 & 78.25 & 81.09 & 254.5 & 885.7 & 61.65 & 65.65 & 78.99 & 179.9 & 451.5\\
    & T3s + Decoder & 74.62 & 78.95 & 81.79 & 242.2 & 857.5 & 61.75 & 65.53 & 79.14 & 181.3 & 461.2 \\
    & T2vec + Decoder & 75.69 & 78.86 & 81.68 & 231.6 & 783.6 & 62.24 & 65.77 & 78.97 & 173.9 & 438.0\\
    & RNTrajRec & \underline{75.80} & \underline{79.35} & \underline{81.86} & \underline{218.5} & \underline{757.0} & \underline{63.39} & \underline{65.84} & \underline{79.25} & \underline{171.3} & \underline{433.9} \\
    & \textbf{MM-STGED (Ours)} & \textbf{78.14} & \textbf{80.06} & \textbf{83.58} & \textbf{197.2} & \textbf{696.0} & \textbf{65.69} & \textbf{66.15} & \textbf{80.74} & \textbf{152.5} & \textbf{400.8}\\
    \midrule
    \multirow{10}{*}{\parbox{2.5cm}{\centering $\mu = 1$ minute $\to \epsilon = 15$ seconds}} & HMM + ShortestPath & 35.92 & 67.92 & 60.16 & 529.7 & 638.2 & 34.75 & 48.46 & 48.67 & 527.2 & 659.3\\
    & Linear + HMM & 68.59 & 65.66 & 66.67 & 707.4 & 1005.2 & 66.17 & 64.72 & 75.22 & 368.3 & 571.1\\
    & MPR + HMM & 62.25 & 62.67 & 60.53 & 418.8 & 659.0 & 66.27 & 65.66 & 74.51 & 402.6 & 628.2\\
    & DHTR + HMM & 51.09 & 63.40 & 50.14 & 584.7 & 750.4 & 52.98 & 69.60 & 57.17 & 420.4 & 625.8 \\
    & AttnMove + Rule & 79.60 & 81.55 & 82.75 & 194.4 & 752.6 & 72.07 & 69.59 & 80.41 & 156.6 & 360.7 \\
    & MTrajRec & 81.12 & 81.73 & 83.75 & 187.1 & 718.4 & 71.65 & 70.92 & 80.35 & 114.9 & 332.3 \\
    
    & T2vec + Decoder & 81.69 & 81.90 & 83.88 & 185.6 & 714.1 & 71.86 & 71.10 & 80.48 & 114.2 & 334.7 \\
    & T3s + Decoder & 80.90 & \underline{82.78} & 83.15 & 187.1 & 713.0 & 71.78 & 71.61 & 80.16 & \underline{110.1} & 328.0\\
    & RNTrajRec & \underline{81.88} & 82.09 & \underline{84.84} & \underline{177.9} & \underline{702.5} & \underline{72.31} & \underline{71.88} & \underline{80.57} & 110.3 & \underline{325.7}\\
    & \textbf{MM-STGED (Ours)} & \textbf{84.26} & \textbf{84.15} & \textbf{85.92} & \textbf{154.0} & \textbf{633.5} & \textbf{73.16} & \textbf{72.27} & \textbf{80.81} & \textbf{108.2} & \textbf{321.9}\\
    \bottomrule
\end{tabular}
}
\label{tab:Porto}
\end{table*}

\subsection{Experiment Results}
Table~\ref{tab:Porto} shows the trajectory recovery results for various sampling intervals of the sparse trajectory on the Chengdu and Porto datasets.
As the sampling intervals increase, the difficulty of recovering the trajectory also rises. The results indicate that our MM-STGED model achieves promising performance compared to all the baselines.

\textcolor{black}{Specifically, when compared to the end-to-end methods, the two-stage methods (HMM + ShortestPath, Linear + HMM, MPR + HMM, and DHTR + HMM) show worse results, especially at large sampling intervals of sparse trajectory. We find that first map-matching and then recovering trajectory point is the worst performance since the accuracy of map-matching is low in the sparse trajectory.
Linear + HMM performs worse than DHTR + HMM. This can be attributed to the fact that DHTR is a deep-learning method and can capture the complex correlations between GPS points, whereas linear interpolation falls short in modeling complex mobility regularity.
MPR + HMM excels in modeling common travel performance by finding popular routes. In comparison to the other two-stage methods, it demonstrates relatively high performance.}

Among the end-to-end solutions, AttnMove only models the correlation of road segments while disregarding the impact of GPS locations, which results in comparatively lower performance.
T2vec uses Bi-LSTM to model trajectory, outperforming MTrajRec and T3s. This demonstrates the effectiveness of taking into account both historical and future temporal dependence among GPS points. 
RNTrajRec further models the spatial dependence between GPS points as well as the neighboring road network. While it reports the highest accuracy among the baselines, its failure to incorporate the semantic information between GPS points limits its performance, rendering it less effective than our proposed model.

Our proposed MM-STGED achieves the best performance. On average, it achieves a 6.09\% improvement on the Chengdu dataset and a 4.24\% improvement on the Porto dataset.
Remarkably, MM-STGED's performance gain is even more obvious for larger sampling intervals. For instance, when the sampling interval $\mu$ of sparse trajectory is set at 4 minutes, compared with the RNTrajRec, the Acc increases from 67.66\% to 70.64\% on the Chengdu dataset and from 54.59\% to 57.30\% on the Porto dataset. Additionally, it reduces the MAE and RMSE by 40 and 56 meters respectively for the Chengdu dataset, and by 25 and 39 meters for the Porto dataset.
These results can be attributed to two key factors. 
Firstly, unlike existing sequence-based models, MM-STGED modeling trajectory from a graph perspective, efficiently integrates the micro-semantic information of the trajectory. 
Secondly, MM-STGED extracts the macro-semantic information between trajectories, which can explicitly model people’s shared travel preferences, subsequently guiding the trajectory recovery process in the right direction.

\begin{table}[h]	
\centering
\caption{Component analysis of MM-STGED.}
\resizebox{1.0\linewidth}{!}{
\begin{tabular}{c|c|ccccc}
	\toprule
	Dataset & Method & Acc($\%$) & Recall($\%$) & Prec($\%$) & MAE & RMSE\\
	\midrule
	\multirow{6}{*}{Chengdu} & w/o GCN & 77.87 & 79.66 & 83.29 & 199.1 & 703.0 \\
	& w/o RN & 77.17 & 78.86 & 82.04 & 212.2 & 736.6\\
	& w/o RC & 77.78 & 79.95 & 83.45 & 205.0 & 721.5\\
	& w/o TLG & 77.29 & 79.59 & 83.52 & 209.3 & 731.2 \\
	& w/o User & 78.04 & 80.02 & 83.50 & 198.0 & 708.4\\
	& \textbf{MM-STGED} & \textbf{78.14} & \textbf{80.06} & \textbf{83.58} & \textbf{197.2} & \textbf{696.0}\\
	\midrule
	\multirow{6}{*}{Porto} & w/o GCN & 65.21 & 65.76 & 80.24 & 156.6 & 412.8\\
	& w/o RN & 64.65 & 65.06 & 80.07 & 161.5 & 417.8 \\
	& w/o RC & 65.18 & 65.50 & 80.03 & 157.5 & 410.6 \\
	& w/o TLG & 65.13 & 65.83 & 79.97 & 156.8 & 407.0 \\
	& w/o User & 65.36 & 65.76 & 80.55 & 156.1 & 405.3 \\
	& \textbf{MM-STGED} & \textbf{65.45} & \textbf{66.28} & \textbf{80.59} & \textbf{153.3} & \textbf{402.7} \\
	\bottomrule
\end{tabular}
}
\label{tab:ablation}
\end{table}
 
\subsection{Ablation Analysis}
To explore the effects of different components, we conduct five ablation experiments under the sampling interval of $\mu = 2 \, \mathrm{minutes} \to \epsilon=15 \, \mathrm{seconds}$. Specifically, we remove some components from our MM-STGED and observe their performance.
\begin{itemize}
	\item \textbf{w/o GCN} removes our graph-based trajectory encoder and uses GRU as the encoder to model the temporal correlation of the GPS points.
	\item \textbf{w/o RN} represents we remove the road network aware layer when encoding the trajectory, and ignore the road network structure information.
	\item \textbf{w/o RC} removes the road condition information in the decoder. This is to prove that extracting the traffic conditions related to the traveling routes is necessary.
	\item \textbf{w/o TLG} removes the macro trajectory flow graph. It represents that the model cannot provide guidance when recovering unobserved GPS points.
	\item \textbf{w/o User} removes the user’s personalized information in the decoder.
\end{itemize}

As shown in Table~\ref{tab:ablation}, we observe that each component in the MM-STGED plays an important role, especially in the capture of the micro- and macro-semantic information. 
Firstly, removing any component that captures micro-semantic information (\textbf{GCN} and \textbf{RN}) will decrease MM-STGED's performance. This proves that it is necessary to consider the sequential information of GPS points and the micro-semantic information when learning trajectory representation.
Secondly, any module in the macro-semantics (\textbf{RC} and \textbf{TLG}) is helpful for improving the performance, since route-related road conditions can provide the environment of the studied trajectory, and the people's travel preferences can provide the prior and narrow the road search space, both of them can guide the trajectory recovery.

\begin{figure}[t]
	\centering
	\includegraphics[width=0.5\textwidth]{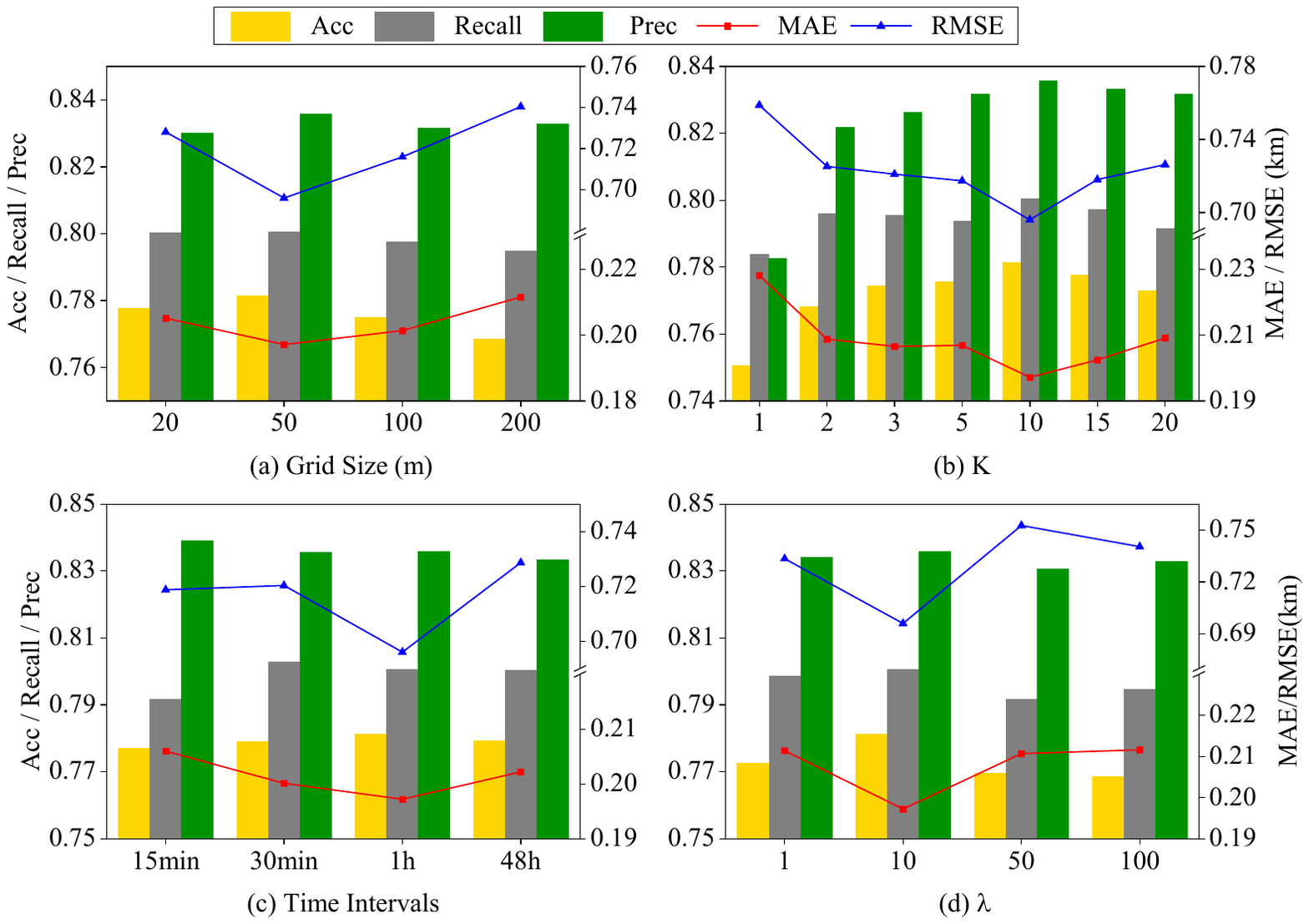}
	\caption{Key hyper-parameters experiment on the Chengdu dataset.}
\label{fig:CD_hyper_parameter}
\end{figure}
\begin{figure}[t]
	\centering
\includegraphics[width=0.5\textwidth]{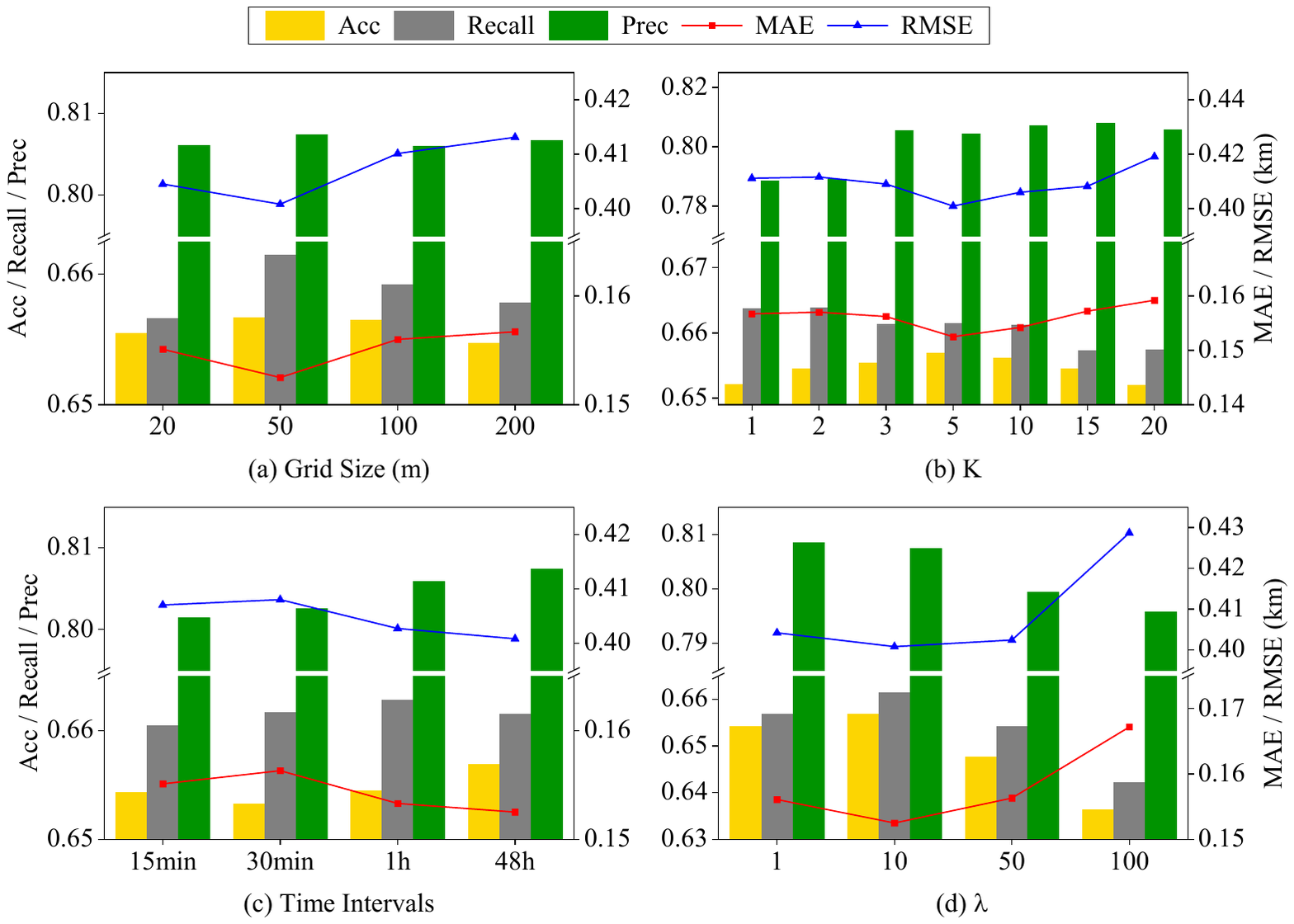}
\caption{Key hyper-parameters experiment on the Porto dataset.}
\label{fig:Porto_hyper_parameter}
\end{figure}
\subsection{Hyper-parameters Experiments}
We further investigate the parameter sensitivity in the case of $\mu = $ 2 minutes $\to$ $\epsilon = $ 15 seconds on the Chengdu and Porto datasets. We evaluate the effectiveness of the grid size divide in the encoder, the top-K highest probability road segment selection in the decoder, the road condition time interval selection, and the weight $\lambda$ of the loss function.

\subsubsection{Grid Size Setting} 
As previously mentioned in the context of graph-based trajectory encoder, we partition the area of interest into grids to facilitate the training reliability. 
The smaller the grid cell size, the higher the accuracy of the captured spatial information, but it will increase the number of grids and introduce the complexity of the model. Similarly, a larger grid cell size simplifies the model at the expense of spatial information accuracy.
Thus, to investigate the trade-off between accuracy and complexity, we set the grid cell size to 20, 50, 100, and 200 meters. As shown in Figure~\ref{fig:CD_hyper_parameter}(a) and \ref{fig:Porto_hyper_parameter}(a), MM-STGED achieves the best performance with a 50-meter grid cell size.

\subsubsection{Top-$K$ Selection} 
To explore the influence of $K$ in the decoder, we select $K$ within the set $\{1, 2, 3, 5, 10, 15, 20\}$. 
As shown in Figure~\ref{fig:CD_hyper_parameter}(b) and \ref{fig:Porto_hyper_parameter}(b), MM-STGED demonstrates optimal performance with $K$ set to 10 for the Chengdu dataset, and 5 for the Porto dataset. For a smaller $K$, there is an accumulation of errors, especially when $K=1$. Conversely, a larger $K$ increase the road segment search space and reduce model efficiency. 
Furthermore, we observe that the Chengdu dataset exhibits significant fluctuations in performance as $K$ varies. To understand the underlying cause, we calculate the average neighbors of road segments in the macro trajectory flow graph. We find that per road has an average of 5.61 neighbors in the Chengdu dataset, whereas the Porto dataset has an average of 27.91. This suggests that the choice of $K$ has a more important impact on the Chengdu dataset in comparison to the Porto dataset.

\subsubsection{Time Interval Setting on the Road Condition} 
During the construction of the Road Condition, the hyper-parameter $T$ represents the temporal period of traffic data considered for extracting the environment in which a given trajectory is located. 
To explore its impact, we investigate the effects of various time interval settings, including 15 minutes, 30 minutes, and 1 hour. In addition, we experiment with an alternative approach whereby a 48-hour time frame is deployed, splitting the data into two distinct 24-hour periods corresponding to weekdays and weekends respectively. 
As shown in Figure~\ref{fig:CD_hyper_parameter}(c) and \ref{fig:Porto_hyper_parameter}(c), MM-STGED attained optimal performance with the time interval set at 1 hour for the Chengdu dataset and 30 minutes for the Porto dataset. As the time interval gets larger, the performance decreases. This can be attributed to the increased sparsity of macro traffic flow derived from historical trajectory data. 

\subsubsection{Loss Function Weight $\lambda$} 
We select loss function weight $\lambda$ within the set $\{1, 10, 50, 100\}$ to analyze its influence on the performance of MM-STGED.
As shown in Figure~\ref{fig:CD_hyper_parameter}(d) and \ref{fig:Porto_hyper_parameter}(d), the performance of MM-STGED achieves the best performance when $\lambda=10$. This suggests that MM-STGED benefits from a good balance of these two tasks.



\subsection{Case Study}

\begin{figure*}
	\centering
		\subfigure{
		\includegraphics[width=0.3\textwidth]{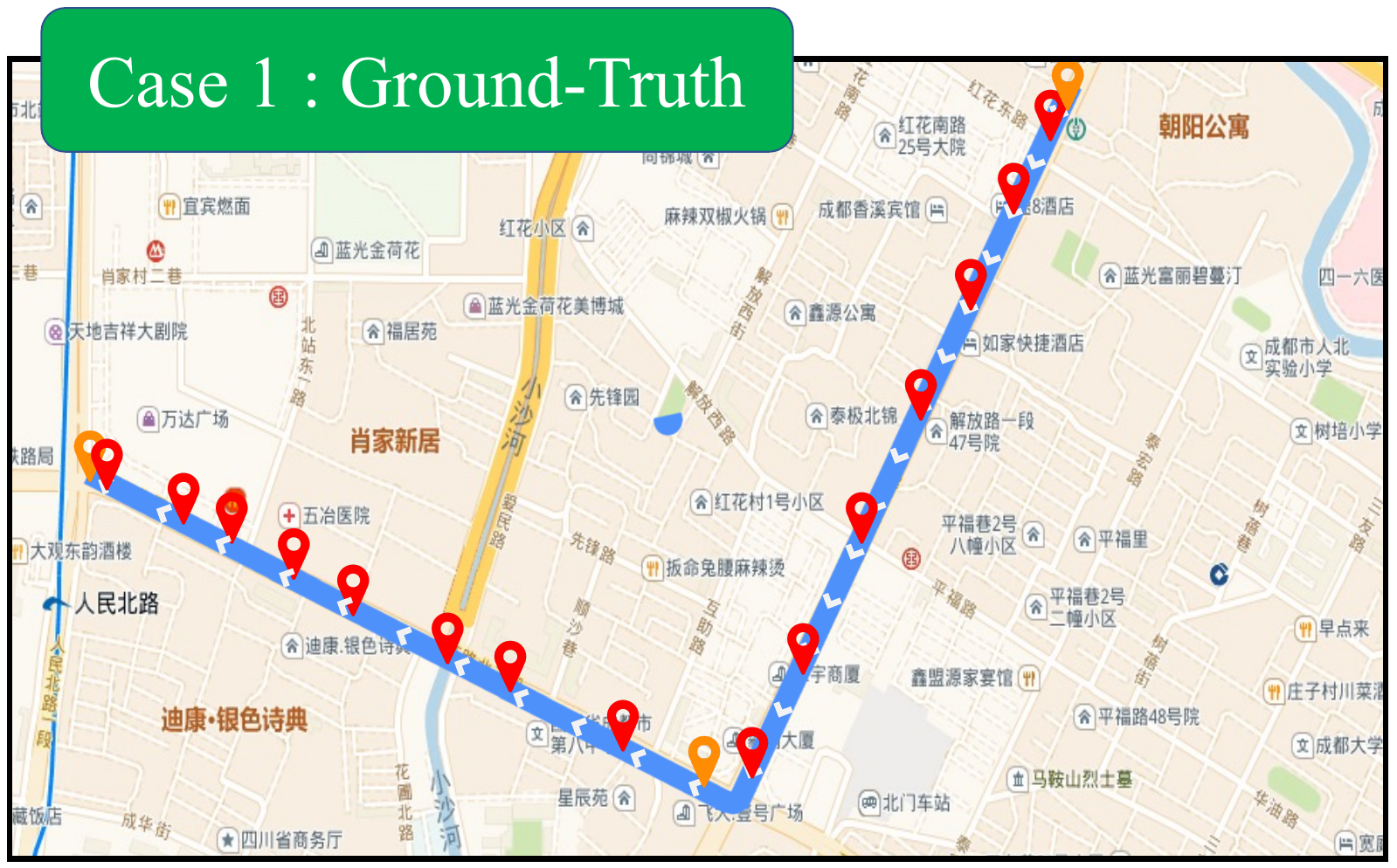}
		}
		\subfigure{
		\includegraphics[width=0.3\textwidth]{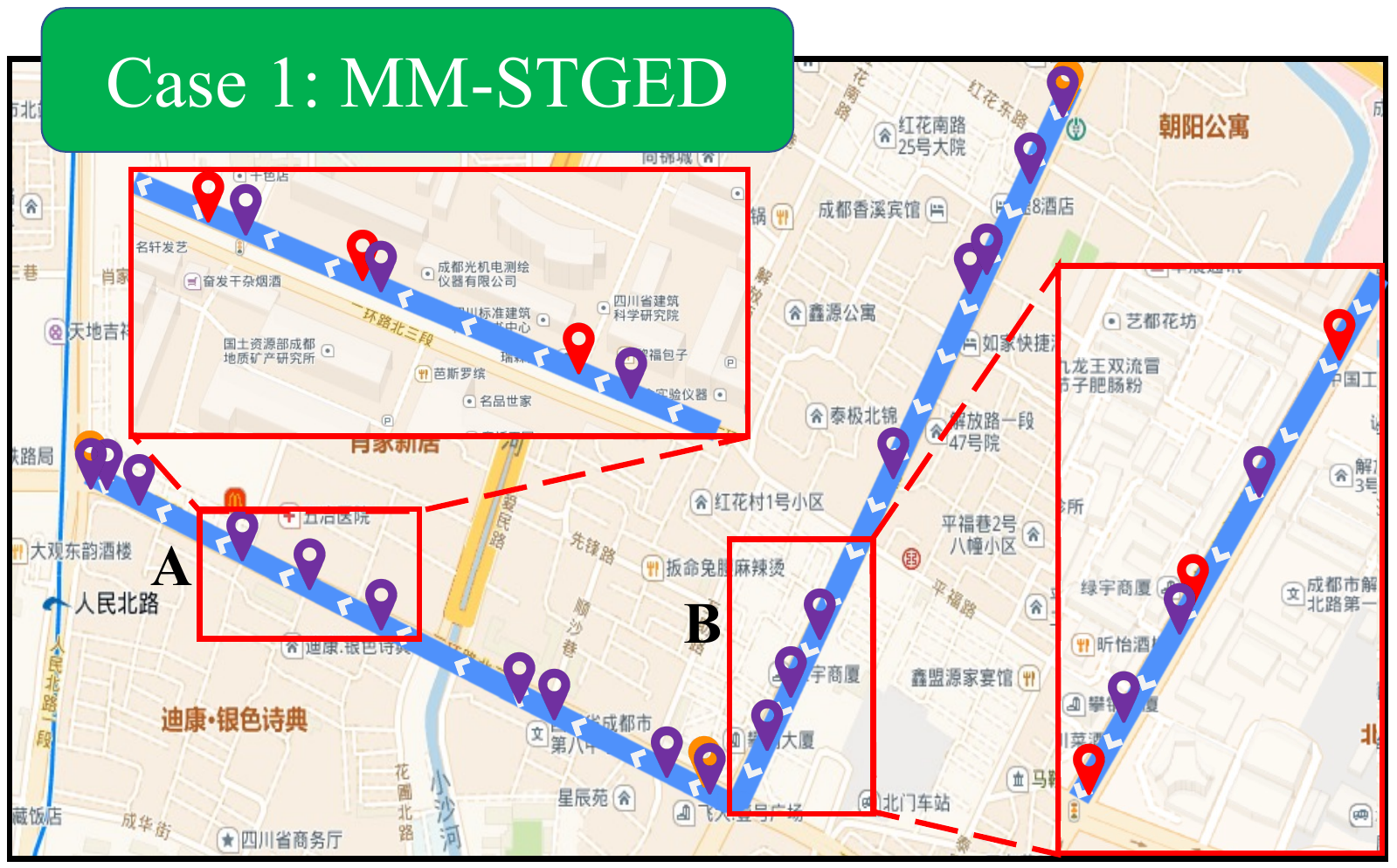}
	}
	\subfigure{
		\includegraphics[width=0.3\textwidth]{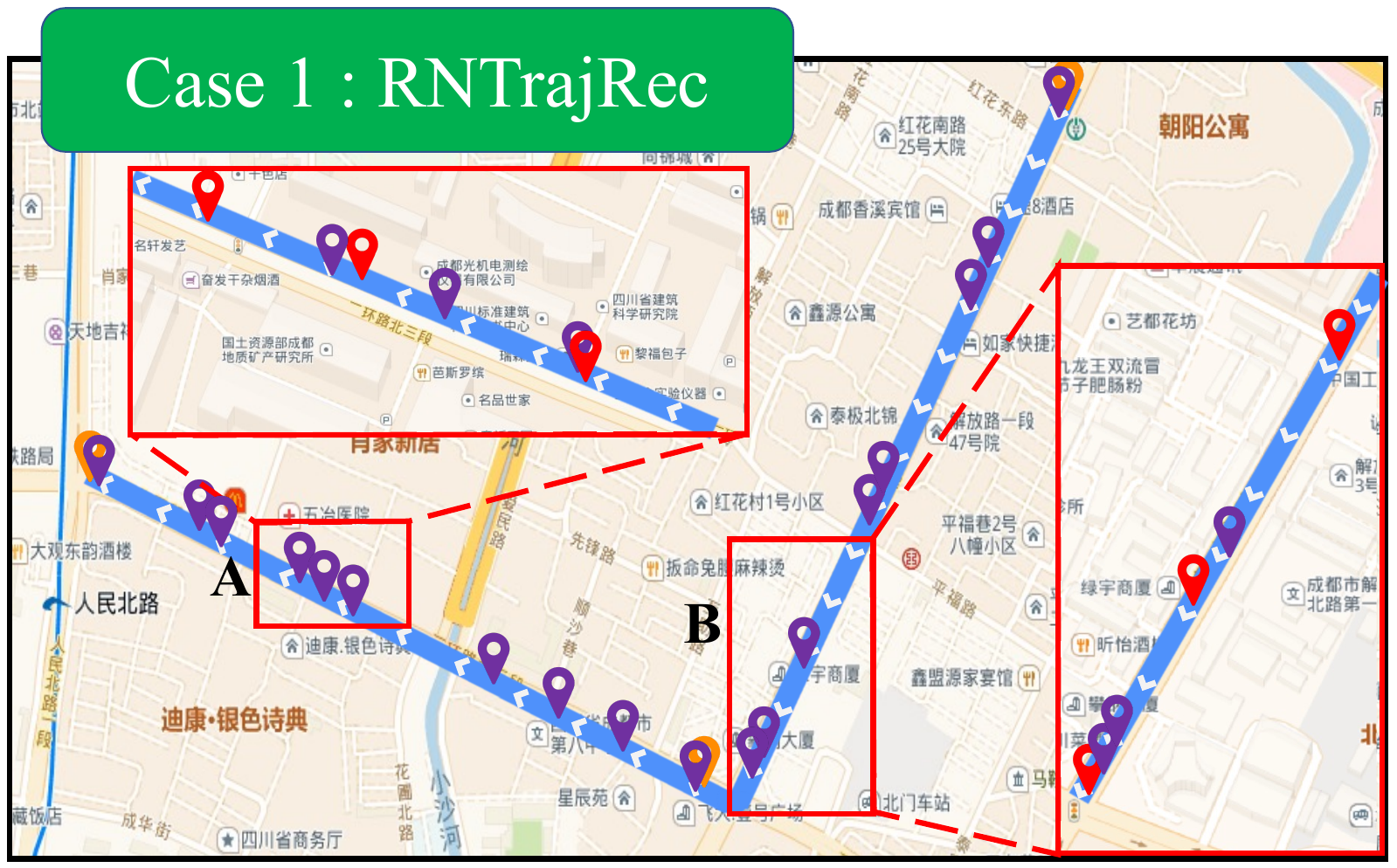}
	}
	\subfigure{
		\includegraphics[width=0.3\textwidth]{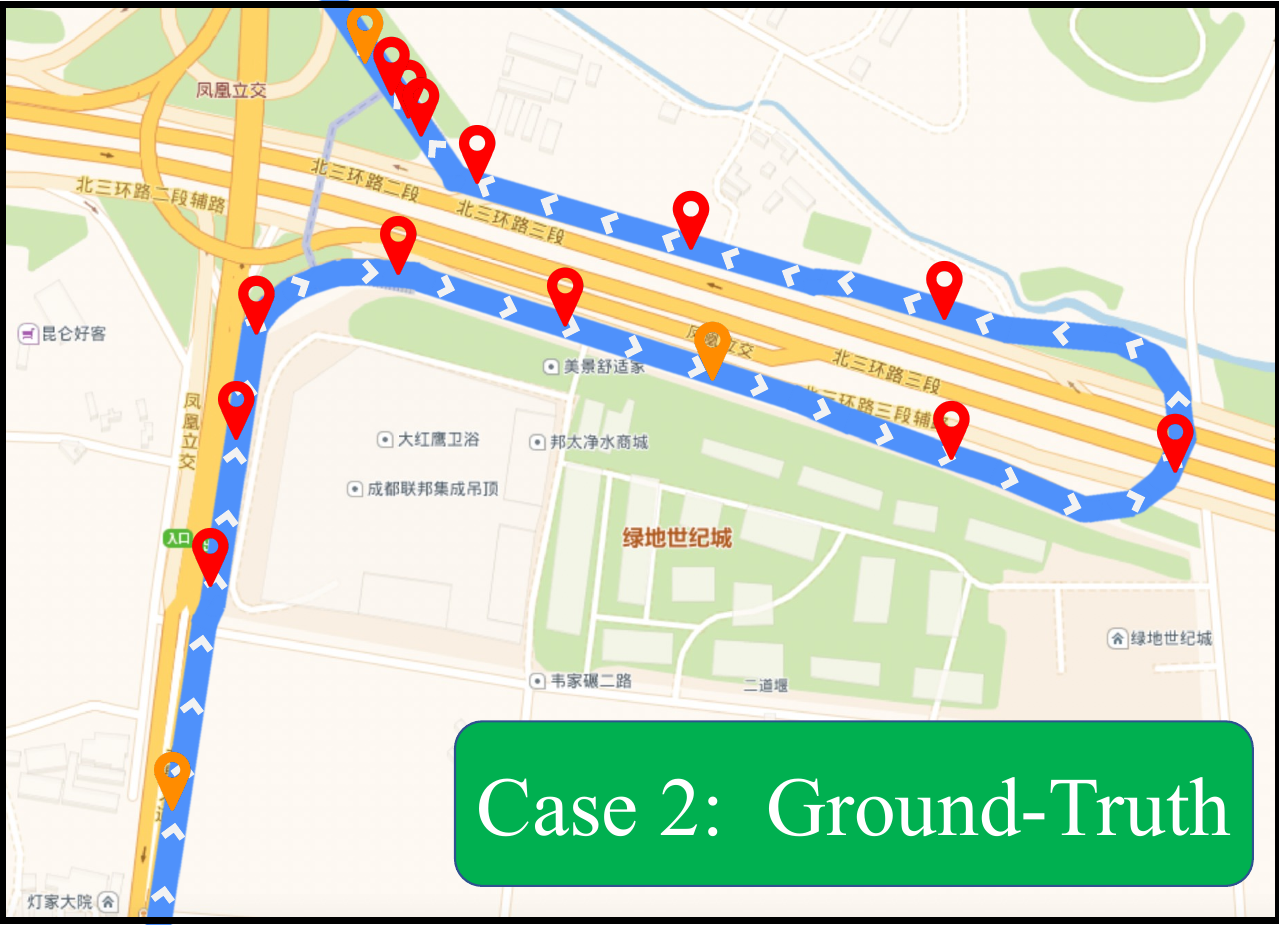}
	}
	\subfigure{
		\includegraphics[width=0.3\textwidth]{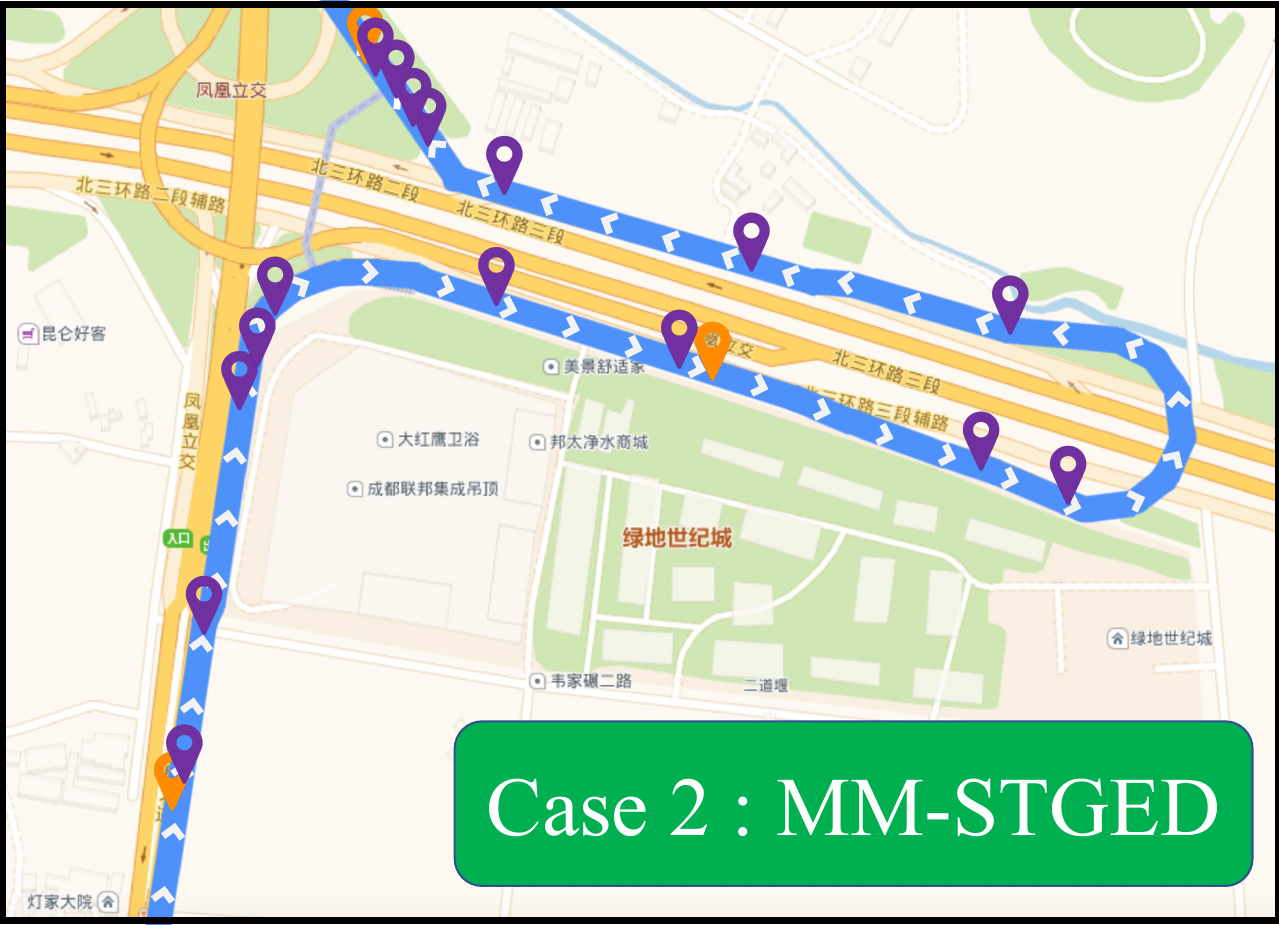}
	}
		\subfigure{
		\includegraphics[width=0.3\textwidth]{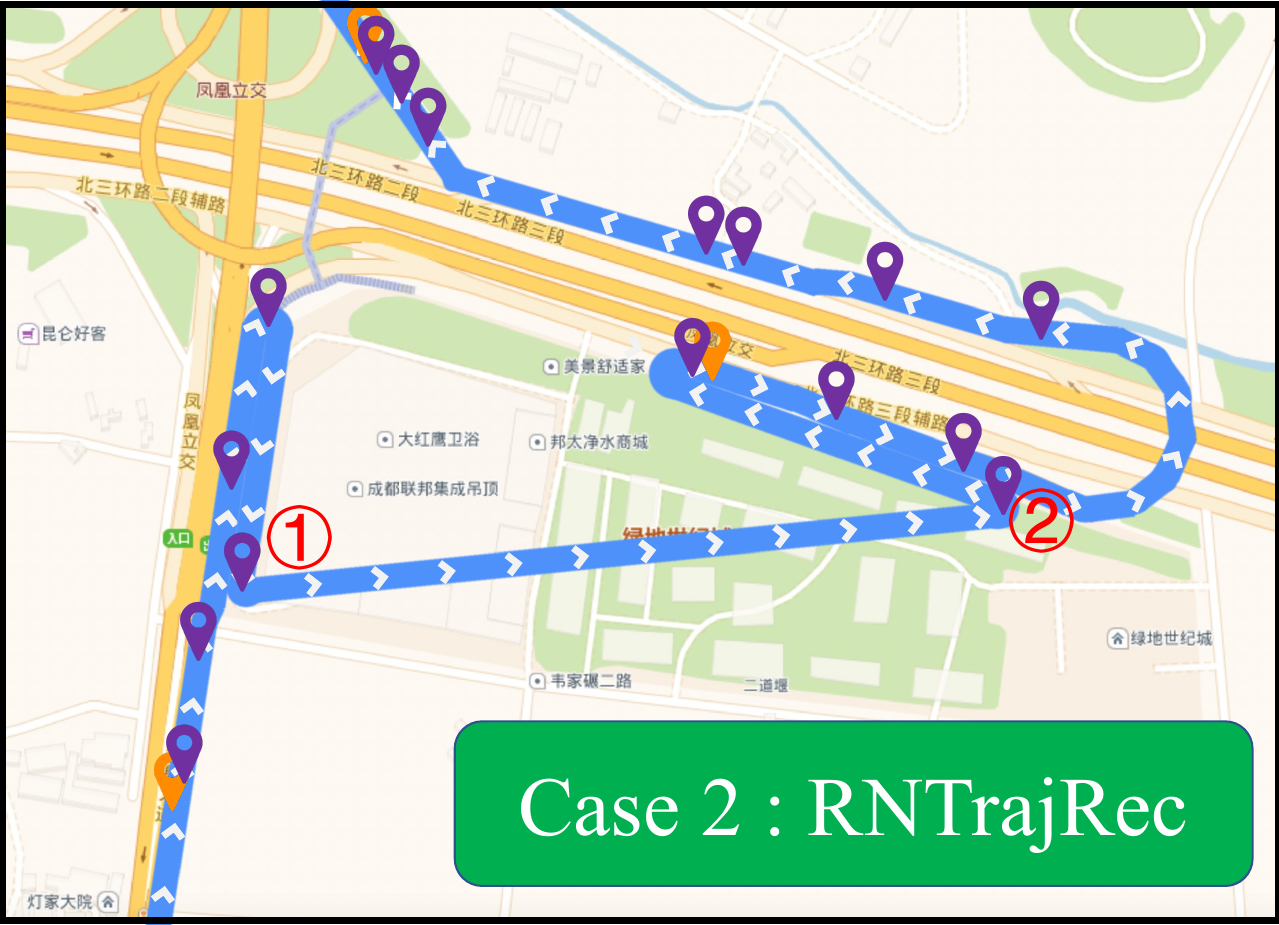}
	}
	
	\caption{Case Study. The blue lines are real route or recovered routes by different models, and orange, red, and purple icons represent the observed, truth, and recovered GPS points, respectively.}
	\label{fig:case_study}
\end{figure*}

To analyze the performance of MM-STGED more intuitively, we provide an empirical case study to visualize the recovery results of our MM-STGED compared with the SOTA baseline model RNTrajRec on the Chengdu dataset.
As shown in Figure~\ref{fig:case_study}, the predicted and truth routes are visually represented by the blue routes, while orange, red, and purple icons represent the observed, truth, and recovered GPS points, respectively. 

In case 1, as shown in the top of Figure~\ref{fig:case_study}, we focus on a simple route for evaluation. Although MM-STGED and RNTrajRec successfully reconstruct the route in alignment with the ground truth, MM-STGED outputs the more precise GPS location.
We further plot the recovered GPS points and the ground truth in two sections (rectangles A and B) in a more intuitive view, and we can observe that our MM-STGED can match the ground truth better.

In case 2, shown in the bottom section of Figure~\ref{fig:case_study}, we select a complex route to visualize the trajectory recovery result.
Here, MM-STGED shows better performance, generating a route that closes the ground truth. In contrast, RNTrajRec outputs an unreasonable route.
This can attribute that RNTrajRec ignores the macro-semantic information for unobserved GPS points, which results in a far distance between two consecutive GPS points and is incapable of formulating a continuous route (the location marked $\textcircled{\scriptsize{1}}$), even leading to detours (at the position labeled $\textcircled{\scriptsize{2}}$). 
While MM-STGED has minor errors between the recovered GPS locations and the ground truth, it still produces a reasonable route. This demonstrates the superiority of macro-semantic information for trajectory recovery for making the right decisions in complex contexts.

\begin{table}[h]	
	\small
	\centering
	\caption{\textcolor{black}{Computation Cost. NA means the model does not require training.}}

	\resizebox{1.02\columnwidth}{!}{
	\begin{tabular}{c|c|c}
		\toprule
		\multirow{2}{*}{Methods} & \multirow{2}{*}{Time Complexity} & Training Time \\ 
            & & (ms / per batch) \\
		\hline
		Linear + HMM & $O(M|\mathcal{R}|^2)$ & NA\\ 
		DHTR + HMM & $O(Nd^2 + Md^2 + M^2 +  M|\mathcal{R}|^2)$ & 79.21 \\
		MTrajRec & $O(Nd^2 + Md^2 + M|\mathcal{R}|)$ & 69.23\\ 
		RNTrajRec & $O(|\mathcal{R}|d^2 + N|\mathcal{R}| + lN^2d + Nd^2 + M |\mathcal{R}|)$ & 98.41 \\
		MM-STGED & $O(N|\mathcal{R}| + Nd^2 + M|\mathcal{R}| + Nd) $ & 87.14\\ 
		
		\bottomrule
	\end{tabular}
}
	\label{tab:cost}
\end{table}

\subsection{Computation Cost}
\textcolor{black}{As shown in Table~\ref{tab:cost}, we analyze the time complexity of the model, where $d$ refers to the number of features, $N$ and $M$ refer to the length of sparsity trajectory and recovered trajectory, $|\mathcal{R}|$ refers to total road segment, $l$ refers to the number of GPSFormer layers in RNTrajRec.}

\textcolor{black}{The time complexity of MM-STGED increases linearly with the number of observed GPS points.
In the graph-based trajectory micro-semantics encoder, the time complexity is $O(Nd^2)$ and $O(N|\mathcal{R}|)$ for the trajectory encoder and road network-aware layer, respectively. 
It's worth noting that extracting macro-semantic information can be performed before training and does not affect the time complexity. 
For the graph-based trajectory recovery decoder, the time complexity is $O(M|\mathcal{R}|)$ and $O(Nd)$ for the road mask layer and dynamic global attention layer, respectively, for every time step. 
Compared to the state-of-the-art method RNTrajRec, our model has a lower time complexity.}

\textcolor{black}{Besides, we compare the training time of MM-STGED with other baselines, in the case of $\mu = 2$ minutes $\to \epsilon = 15$ seconds on the Chengdu dataset. Linear + HMM does not require training because it is rule-based. MM-STGED takes around 87.14ms per batch, with a batch size of 128 using an NVIDIA RTX A4000 card. Compared with baselines, despite the training time of MM-STGED not being the fastest, it is still acceptable since the fastest model, MTrajRec, is implemented by GRU, and the performance of our MM-STGED outperforms MTrajRec by about 3.62\% in Acc and 57.2 meters in MAE.}

\section{Conclusion}
In this paper, we propose MM-STGED, an end-to-end encoder-decoder method for map-constrained trajectory recovery. 
MM-STGED models trajectories from a graph perspective to describe the micro-semantic information, simultaneously capturing the explicit absolute information of each GPS point and their implicit relative information. And a novel graph convolution is proposed that simultaneously aggregates nodes and edges information together.
Furthermore, the macro-semantic information reflected by a group of trajectories is extracted, followed by a well-designed graph-based decoder to efficiently recover trajectory points. Experimental results demonstrate MM-STGED's superior performance compared to the state-of-the-art baselines. 

\section*{Acknowledgments}
This work was supported by the National Natural Science Foundation of China (Grant No. 62202043) and the Beijing Natural Science Foundation (Grant No. 4242029).

\bibliographystyle{IEEEtran}
\bibliography{main}


\end{document}